\documentclass[10pt,twocolumn,letterpaper]{article}

\usepackage{cvpr}
\usepackage{times}
\usepackage{epsfig}
\usepackage{graphicx}
\usepackage{amsmath}
\usepackage{amssymb}
\usepackage{multirow}
\usepackage{booktabs}
\usepackage[]{algorithm2e}
\usepackage{caption,subcaption}

\DeclareMathOperator*{\argmin}{arg\,min}
\newtheorem{theorem}{Theorem}

\usepackage[pagebackref=true,breaklinks=true,letterpaper=true,colorlinks,bookmarks=false]{hyperref}
\usepackage{ifpdf} 

\cvprfinalcopy 

\def\cvprPaperID{6101} 

\begin{document}

\title{Group Sparsity: The Hinge Between Filter Pruning and Decomposition for Network Compression}

\author{Yawei Li$^1$, Shuhang Gu$^1$, Christoph Mayer$^1$,  Luc Van Gool$^{1, 2}$, Radu Timofte$^1$\\
$^1$Computer Vision Lab, ETH Z\"urich, Switzerland, $^2$KU Leuven, Belgium\\
{\tt\small \{yawei.li, shuhang.gu, chmayer, vangool, radu.timofte\}@vision.ee.ethz.ch}
}

\maketitle

\begin{abstract}
In this paper, we analyze two popular network compression techniques, \ie filter pruning and low-rank decomposition, in a unified sense. By simply changing the way the sparsity regularization is enforced, filter pruning and low-rank decomposition can be derived accordingly. This provides another flexible choice for network compression because the techniques complement each other. For example, in popular network architectures with shortcut connections (\eg ResNet), filter pruning cannot deal with the last convolutional layer in a ResBlock while the low-rank decomposition methods can. In addition, we propose to compress the whole network jointly instead of in a layer-wise manner. Our approach proves its potential as it compares favorably to the state-of-the-art on several benchmarks. Code is available at \url{https://github.com/ofsoundof/group_sparsity}.
\end{abstract}

\section{Introduction}

\begin{figure*}[t]
    \begin{center}
        \includegraphics[width=1.0\linewidth]{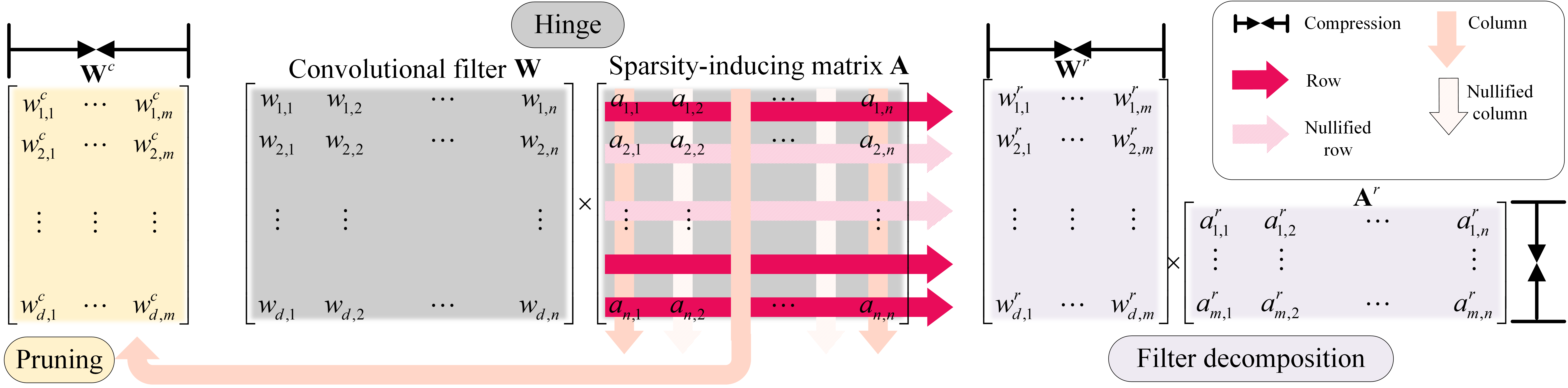}
    \end{center}
    \vspace{-0.4cm}
    \caption{A sparsity-inducing matrix $\mathbf{A}$ is attached to a normal convolution. The matrix acts as the hinge between filter pruning and decomposition. By enforcing group sparsity to the columns and rows of the matrix, equivalent pruning and decomposition operations can be obtained. For pruning, the product of $\mathbf{W}$ and the column-reduced matrix $\mathbf{A}^c$, \ie $\mathbf{W}^c$ acts as the new convolutional filter. To save computation during decomposition the reduced matrices $\mathbf{W}^r$ and $\mathbf{A}^r$ are used as two convolutional filters.}
    \vspace{-0.4cm}
    \label{fig:the_hinge_between_pruning_decomposition}
\end{figure*}

During the past years, convolutional neural networks (CNNs) have reached state-of-the-art performance in a variety of computer vision tasks~\cite{krizhevsky2012imagenet,he2016deep,huang2017densely,long2015fully,danelljan2017eco,redmon2016you,zhang2017beyond,li2019_3dappearance}. 
However, millions of parameters and heavy computational burdens are indispensable for new advances in this field.
This is not practical for the deployment of neural network solutions on edge devices and mobile devices.

To overcome this problem, neural network compression emerges as a promising solution, aiming at a lightweight and efficient version of the original model.
Among the various network compression methods, filter pruning and filter decomposition (also termed low-rank approximation) have been developing steadily.
Filter pruning nullifies the weak filter connections that have the least influence on the accuracy of the network while low-rank decomposition converts a heavy convolution to a lightweight one and a linear combination~\cite{he2017channel,He_2018_ECCV_AMC,li2019learning}.   
Despite their success, both the pruning-based and decomposition-based approaches have their respective limitations.
Filter pruning can only take effect in pruning output channels of a tensor and equivalently cancelling out inactive filters. 
This is not feasible under some circumstances. 
The skip connection in a block is such a case where the output feature map of the block is added to the input. 
Thus, pruning the output could amount to cancelling a possible important input feature map. 
This is the reason why many pruning methods fail to deal with the second convolution of the ResNet~\cite{he2016deep} basic block. 
As for filter decomposition, it always introduces another $1 \times 1$ convolutional layer, which means additional overhead of calling CUDA kernels.

Previously, filter pruning and decomposition were developed separately. 
In this paper, we unveil the fact that filter pruning and decomposition are highly related from the viewpoint of compact tensor approximation. %
Specifically, both filter pruning and filter decomposition seek a compact approximation of the parameter tensors despite their different operation forms to cope with the application scenarios.
Consider a vectorized image patch $\mathbf{x} \in \mathbb{R}^{m\times1}$ and a group of $n$ filters $\mathbf{W}=\{\mathbf{w}_1, \dots, \mathbf{w}_n\} \in  \mathbb{R}^{m\times n}$.
The pruning methods remove output channels and approximate the original output $\mathbf{x}^T \mathbf{W}$ as $\mathbf{x}^T \mathbf{C}$, where $\mathbf{C} \in \mathbb{R}^{m\times k}$ only has $k$ output channels.
Filter decomposition methods approximate $\mathbf{W}$ as two filters $\mathbf{A} \in \mathbb{R} ^ {m\times k}$ and $\mathbf{B} \in \mathbb{R} ^ {k\times n}$ and $\mathbf{A}\mathbf{B}$ is the rank $k$ approximation of $\mathbf{W}$.
Thus, both the pruning and decomposition based methods seek a compact approximation to the original network parameters, but adopt different strategies for the approximation.

The above observation shows that filter pruning and decomposition constitute complementary components of each other. This fact encourages us to design a unified framework that is able to incorporate the pruning-based and decomposition-based approaches simultaneously.
This simple yet effective measure can endow the devised algorithm with the ability of flexibly switching between the two operation modes, \ie filter pruning and decomposition, depending on the layer-wise configurations.
This makes it possible to leverage the benefits of both methods.

The hinge point between pruning and decomposition is group sparsity, see Fig.~\ref{fig:the_hinge_between_pruning_decomposition}. 
Consider a 4D convolutional filter, reshaped into a 2D matrix $\mathbf{W} \in \mathbb{R}^{\textrm{features} \times \textrm{outputs}}$. 
Group sparsity is added by introducing a sparsity-inducing matrix $\mathbf{A}$. 
By applying group sparsity constraints on the columns of $\mathbf{A}$, the output channel of the sparsity-inducing matrix $\mathbf{A}$ and equivalently of the matrix product $\mathbf{W} \times \mathbf{A}$ can be reduced by solving an optimization problem. 
This is equivalent to filter pruning. 
On the other hand, if the group sparsity constraints are applied on the rows of $\mathbf{A}$, then the inner channels of the matrix product $\mathbf{W} \times \mathbf{A}$, namely, the output channel of $\mathbf{W}$ and the input channel of $\mathbf{A}$, can be reduced. 
To save the computation, the single heavyweight convolution $\mathbf{W}$ is converted to a lightweight and a $1 \times 1$ convolution with respect to the already reduced matrices $\mathbf{W}^r$ and $\mathbf{A}^r$. 
This breaks down to filter decomposition.

Thus, the contribution of this paper is four-fold. 
\begin{enumerate}
    \item[I] Starting from the perspective of \textbf{\textit{compact tensor approximation}}, the connection between filter pruning and decomposition is analyzed. Although this perspective is the core of filter decomposition, it is still novel for network pruning. Actually, both of the methods approximate the weight tensor with compact representation that keeps the accuracy of the network. 
    
    \item[II] Based on the analysis, we propose to use \textbf{\textit{sparsity-inducing matrices}} to hinge filter pruning and decomposition and bring them under the same formulation. This square matrix is inspired by filter decomposition and corresponds to a $1 \times 1$ convolution. By changing the way how the sparsity regularizer is applied to the matrix, our algorithm can achieve equivalent effect of either filter pruning or decomposition or both. To the best of our knowledge, this is the first work that tries to analyze the two methods under the same umbrella.
    
    \item[III] The third contribution is the developed \textbf{\textit{binary search, gradient based learning rate adjustment, layer balancing, and annealing methods}} that are important for the success of the proposed algorithm. Those details are obtained by observing the influence of the proximal gradient method on the filter during the optimization. 
    
    \item[IV] The proposed method can be \textbf{\textit{applied to various CNNs}}. We apply this method to VGG~\cite{simonyan2014very}, ResNet~\cite{he2016deep}, ResNeXt~\cite{xie2017aggregated}, WRN~\cite{zagoruyko2016wide}, and DenseNet~\cite{huang2017densely}. The proposed network compression method achieves state-of-the-art performance on those networks. 

\end{enumerate}

The rest of the paper is organized as follows. Sec.~\ref{sec:related_works} discusses the related work. Sec.~\ref{sec:method} explains the proposed network compression method. Sec.~\ref{sec:implementation} describes the implementation considerations. The experimental results are shown in Sec.~\ref{sec:experimental_results}. Sec.~\ref{sec:conclusion} concludes this paper.

\section{Related Work}
\label{sec:related_works}
In this section, we firstly review the closely related work including decomposition-based and pruning-based compression methods.
Then, we list other categories of network compression works.

\subsection{Parameter Pruning for Network Compression}
\textbf{Non-structural pruning.} To compress neural networks,
network pruning disables the weak connections in a network that have a small influence on its prediction accuracy. 
Earlier pruning methods explore unstructured network weight pruning by deactivating connections corresponding to small weights or by applying sparsity regularization to the weight parameters~\cite{han2015deep,liu2015sparse,han2015learning}. 
The resulting irregular weight parameters of the network are not implementation-friendly, which hinders the real acceleration rate of the pruned network over the original one. 

\textbf{Structural pruning.}
To circumvent the above problem, structural pruning approaches zero out structured groups of the convolutional filters~\cite{he2017channel,he2019filter}.
Specifically, group sparsity regularization has been investigated in recent works for the structural pruning of network parameters~\cite{zhou2016less,wen2016learning,alvarez2016learning}.
Wen~\etal~\cite{wen2016learning} and Alvarez~\etal~\cite{alvarez2016learning} proposed to impose group sparsity regularization on network parameters to reduce the number of feature map channels in each layer.
The success of this method triggered the studies of group sparsity based network pruning.
Subsequent works improved group sparsity based approaches in different ways.
One branch of works combined the group sparsity regularizer with other regularizers for network pruning.
A low-rank regularizer \cite{alvarez2017compression} as well as an exclusive sparsity regularizer \cite{yoon2017combined} were adopted for improving the pruning performance.
Another branch of research investigated a better group-sparsity regularizer for parameter pruning including group ordered weighted $\ell_1$ regularizer \cite{zhang2018learning}, 
out-in-channel sparsity regularization \cite{li2019oicsr} and guided attention for sparsity learning \cite{torfi2019gasl}.
In addition, some works also attempted to achieve group-sparse parameters in an indirect manner.
In \cite{Liu_2017_ICCV_Learning} and \cite{huang2018data}, scaling factors were introduced to scale the outputs of specific structures or feature map channels to structurally prune network parameters. 

\begin{figure}[t]
    \begin{center}
        \includegraphics[width=1.0\linewidth]{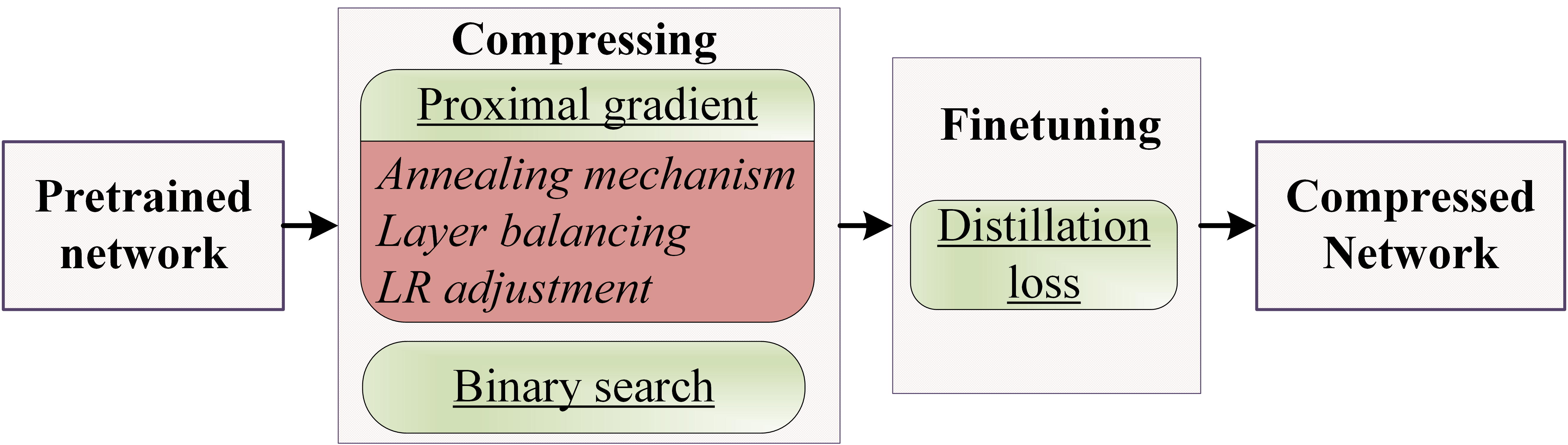}
    \end{center}
    \vspace{-0.4cm}
    \caption{The flowchart of the proposed algorithm.}
    \vspace{-0.4cm}
    \label{fig:flowchart}
\end{figure}

\subsection{Filter Decomposition for Network Compression}
Another category of works compresses network parameters through tensor decomposition.
Specifically, the original filter is decomposed into a lightweight one and a linear projection which contain much fewer parameters than the original, thus resulting in the reduction of parameters and computations.
Early works apply matrix decomposition methods such as SVD~\cite{denton2014exploiting} or CP-decomposition~\cite{lebedev2015speeding}
to decompose 2D filters.
In~\cite{jaderberg2014speeding}, 
Jaderberg~\etal proposed to
approximate the 2D filter set by a linear combination of a
smaller basis set of 2D separable filters.
Subsequent filter basis decomposition works polished the approach in~\cite{jaderberg2014speeding} by using a shared filter basis \cite{son2018clustering} or by enabling more flexible filter decomposition.
In addition, Zhang \etal \cite{zhang2016accelerating} took the input channel as the third dimension and directly compress the 3D parameter tensor.

\subsection{Other methods}

Other network compression methods include network quatization and knowledge distillation. 
Network quantization aims at a low-bit representation of network parameters to save storage and to accelerate inference. 
This method does not change the architecture of the fully-fledged network~\cite{wang2019haq,nagel2019data}. 
Knowledge distillation transfers the knowledge of a teacher network to a student network~\cite{hinton2015distilling}.
Current research in this direction focuses on the architectural design of the student network~\cite{crowley2018moonshine,ashok2018n2n} and the loss function~\cite{tung2019similarity}.

\section{The proposed method}
\label{sec:method}

\begin{figure*}
\begin{minipage}[c]{0.5\textwidth}
  \vspace*{\fill}
  \centering
  \includegraphics[width=0.9\linewidth]{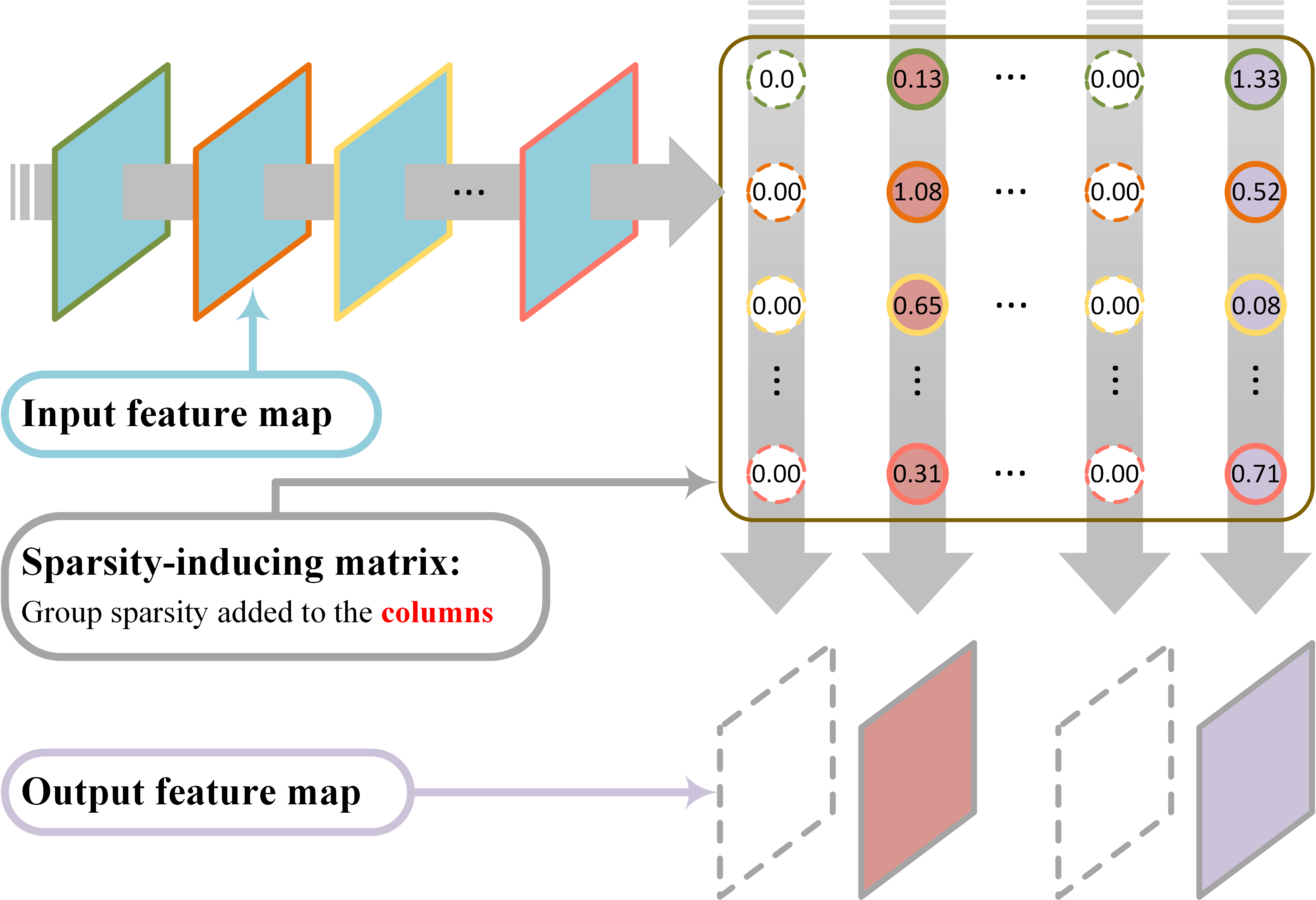}
  \subcaption{Group sparsity enforced on column.}
  \vspace{-0.2cm}
  \label{fig:group_sparsity_column}
\end{minipage}%
\begin{minipage}[c]{0.5\textwidth}
  \vspace*{\fill}
  \centering
  \includegraphics[width=0.9\linewidth]{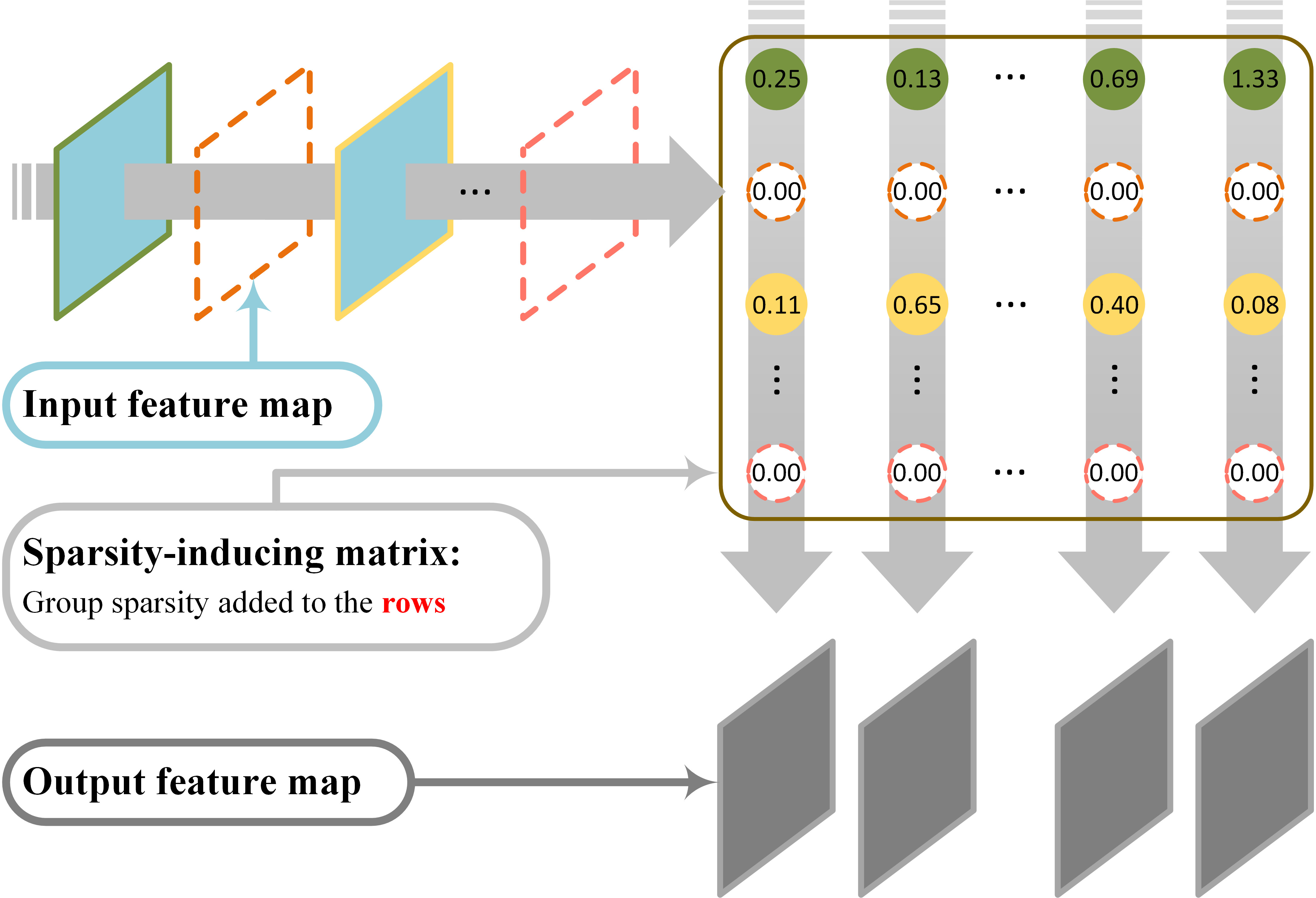}
  \subcaption{Group sparsity enforced on row.}
  \vspace{-0.2cm}
  \label{fig:group_sparsity_row}
\end{minipage}%
\caption{Group-sparsity regularization enforcement: (a) the columns of the sparsity-inducing matrix are regularized. This results in nullified filters and the corresponding output feature maps are removed. (b) the rows are regularized and some are zeroed out. The filters of the previous layer and also the feature maps are removed.}
\vspace{-0.4cm}
\label{fig:group_sparsity}
\end{figure*}

This section explains the proposed method (Fig.~\ref{fig:flowchart}). Specifically, it describes how group sparsity can hinge filter pruning and decomposition. The pair $\{\mathbf{x}, \mathbf{y}\}$ denotes the input and target of the network. Without loss of clarity, we also use $\mathbf{x}$ to denote the input feature map of a layer. The output feature map of a layer is denoted by $\mathbf{z}$. The filters of a convolutional layer are denoted by $\mathbf{W}$ while the introduced group sparsity matrix is denoted by $\mathbf{A}$. The rows and columns of $\mathbf{A}$ are denoted by $\mathbf{A}_i$, and $\mathbf{A}_j$, respectively. The general structured groups of $\mathbf{A}$ are denoted by $\mathbf{A}_g$.

\subsection{Group sparsity}
\label{subsec:group_sparsity}

The convolution between the input feature map $\mathbf{x}$ and the filters can be converted to a matrix multiplication, \ie,
\begin{equation}
    \mathbf{Z} = \mathbf{X} \times \mathbf{W},
\label{eqn:convolution}
\end{equation}
where $\mathbf{X} \in\mathbb{R}^{N \times cwh}$,  $\mathbf{W} \in\mathbb{R}^{cwh \times n}$, and $\mathbf{Z} \in\mathbb{R}^{N \times n}$ are the reshaped input feature map, output feature map, and convolutional filter, $c$, $n$, $w \times h$, and $N$ denotes the input channel, number of filters, filter size, and number of reshaped features, respectively. For the sake of brevity, the bias term is omitted here. The weight parameters $\mathbf{W}$ are usually trained with some regularization such as weight decay to avoid overfitting the network. To get structured pruning of the filter, structured sparsity regularization is used to constrain the filter, \ie
\begin{equation}
    \min_{\mathbf{W}}\mathcal{L}(\mathbf{y}, f(\mathbf{x}; \mathbf{W})) + \mu \mathcal{D}(\mathbf{W}) + \lambda \mathcal{R}(\mathbf{W}),
\end{equation}
where $\mathcal{D}(\cdot)$ and $\mathcal{R}(\cdot)$ are the weight decay and sparsity regularization, $\mu$ and $\lambda$ are the regularization factors.

Different from other group sparsity methods that directly regularize the matrix $\mathbf{W}$~\cite{yoon2017combined,li2019oicsr}, we enforce group sparsity constraints by incorporating a sparsity-inducing matrix $\mathbf{A} \in\mathbb{R}^{n \times n}$, which can be converted to the filter of a $1 \times 1$ convolutional layer after the original layer. Then the original convolution in Eqn.~\ref{eqn:convolution} becomes $\mathbf{Z} = \mathbf{X} \times (\mathbf{W} \times \mathbf{A})$. 
To obtain a structured sparse matrix, group sparsity regularization is enforced on $\mathbf{A}$. Thus, the loss function becomes
\begin{equation}
    \min_{\mathbf{W},\mathbf{A}}\mathcal{L}(\mathbf{y}, f(\mathbf{x}; \mathbf{W}, \mathbf{A})) +  \mu \mathcal{D}(\mathbf{W}) + \lambda \mathcal{R}(\mathbf{A}).
    \label{eqn:group_sparsity}
\end{equation}
Solving the problem in Eqn.~\ref{eqn:group_sparsity} results in structured group sparsity in matrix $\mathbf{A}$. By considering matrix $\mathbf{W}$ and $\mathbf{A}$ together, the actual effect is that the original convolutional filter is compressed.

In comparison with the filter selection method~\cite{Liu_2017_ICCV_Learning,huang2018data}, the proposed method not only selects the filters in a layer, but also makes linear combinations of the filters to minimize the error between the original and the compact filter. On the other hand, different from other group sparsity constraints~\cite{yoon2017combined,li2019oicsr}, there is no need to change the original filters $\mathbf{W}$ of the network too much during optimization of the sparsity problem. In our experiments, we set a much smaller learning rate for the pretrained weight $\mathbf{W}$.

\subsection{The hinge}
The group sparsity term in Eqn.~\ref{eqn:group_sparsity} controls how the network is compressed. This term has the form 
\begin{equation}
    \mathcal{R}(\mathbf{A}) = \Phi ( \|\mathbf{A}_{g}\|_2),
    \label{eqn:sparsity_regularier}
\end{equation}
where $\mathbf{A}_g$ denotes the different groups of $\mathbf{A}$, $\|\mathbf{A}_{g}\|_2$ is the $\ell_2$ norm of the group, and $\Phi(\cdot)$ is a function of the group $\ell_2$ norms.

If group sparsity regularization is added to the columns of $\mathbf{A}$ as in Fig.~\ref{fig:group_sparsity_column}, \ie, $\mathcal{R}(\mathbf{A}) = \Phi ( \|\mathbf{A}_{j}\|_2)$, a column pruned version $\mathbf{A}^c$ is obtained and the output channels of the corresponding $1 \times 1$ convolution are pruned. In this case, we can multiply $\mathbf{W}$ and $\mathbf{A}^c$ and use the result as the filter of the convolutional layer. This is equivalent to pruning the output channels of the convolutional layer with the filter $\mathbf{W}$.

On the other hand, group sparsity can be also applied to the rows of $\mathbf{A}$, \ie $\mathcal{R}(\mathbf{A}) = \Phi (\|\mathbf{A}_{i}\|_2)$. In this case, a row-sparse matrix $\mathbf{A}^r$ is derived and the input channels of the $1 \times 1$ convolution can be pruned (See Fig.~\ref{fig:group_sparsity_row}). Accordingly, the corresponding output channels of the former convolution with filter $\mathbf{W}$ can be also pruned. However, since the output channel of the later convolution is not changed, multiplying out the two compression matrices do not save any computation. So a better choice is to leave them as two separate convolutional layers. This tensor manipulation method is equivalent to filter decomposition where a single convolution is decomposed into a lightweight one and a linear combination. In conclusion, by enforcing group sparsity to the columns and rows of the introduced matrix $\mathbf{A}$, we can derive two tensor manipulation methods that are equivalent to the operation of filter pruning and decomposition, respectively. This provides a degree of freedom to choose the tensor manipulation method depending on the specifics of the underlying network.

\subsection{Proximal gradient solver}

\RestyleAlgo{boxruled}
\begin{algorithm}[h]
    \KwData{training dataset}
    \KwResult{the compressed network}
    initialization: the current compression ratio $\gamma_c = 1$; the target compression ratio $\gamma^\ast$, the nullifying threshold of the group $\ell_2$ norm $\mathcal{T}$\;
    \While{$\gamma_c - \gamma^\ast <= \alpha$}
    {
        start a a new epoch\;
        \For{batch $\in$ training dataset}
        {
            $\mathbf{W}_{t+1} = \mathbf{W}_{t} - \eta_s \nabla \mathcal{G}(\mathbf{W}_{t})$\;
            $\mathbf{A}_{t+\Delta} = \mathbf{A}_{t} - \eta \nabla\mathcal{H}(\mathbf{A}_{t})$\;
            $ \mathbf{A}_{t+1} = \mathbf{prox}_{\lambda\eta\mathcal{R}}(\mathbf{A}_{t+\Delta})$\;
        }
        compress the network with the threshold $\mathcal{T}$\;
        compute the compression ratio $\gamma_c$
    }
    \caption{The optimization algorithm to solve the problem defined in Eqn.~\ref{eqn:group_sparsity}.}
    \label{alg:proximal_gradient}
    \vspace{-0.2cm}
\end{algorithm}
\begin{algorithm}
    \KwResult{the nullifying threshold $\mathcal{T}^\ast = g^{-1}(\gamma^\ast)$}
    initialization: the target compression ratio $\gamma^\ast$, the initial step $s$, the stop criterion $\mathcal{C}$, and $\mathcal{T} = \mathcal{T}_0$\;
    \While{ $|\gamma_n - \gamma^\ast| > \mathcal{C}$}{
        compress the network with the threshold $\mathcal{T}$\;
        calculate the current compression ratio $\gamma_{n}$\;
        \lIf{$(\gamma_{n-1} >= \gamma^\ast) == (\gamma_{n} < \gamma^\ast)$}{$s \gets s / 2$}
        \eIf{$\gamma_n > \gamma_t$}{
            $\mathcal{T} \gets \mathcal{T} + s$\;
        }{
            $\mathcal{T} \gets \mathcal{T} - s$\;
        }
        
    }
    \caption{Binary search of the threshold $\mathcal{T}$.}
    \label{alg:binary_search}
    \vspace{-0.2cm}
\end{algorithm}
To solve the problem defined by Eqn.~\ref{eqn:group_sparsity}, the parameter $\mathbf{W}$ can be updated with stochastic gradient descent (SGD) but with a small learning rate, \ie $\mathbf{W}_{t+1} = \mathbf{W}_{t} - \eta_s \nabla \mathcal{G}(\mathbf{W}_{t})$, where $\mathcal{G}(\mathbf{W}_{t}) = \mathcal{L}(\cdot\,, f(\cdot\,; \mathbf{W}_{t}, \cdot)) + \mu \mathcal{D}(\mathbf{W}_{t})$. This is because it is not desired to modify the pretrained parameters too much during the optimization phase. The focus should be the sparsity matrix $\mathbf{A}$.

The proximal gradient algorithm~\cite{parikh2014proximal} is used to optimize the matrix $\mathbf{A}$ in Eqn.~\ref{eqn:group_sparsity}. It consists of two steps, \ie the gradient descent step and the proximal step. The parameters in $\mathbf{A}$ are first updated by SGD with the gradient of the loss function $\mathcal{H}(\mathbf{A}) = \mathcal{L}(\mathbf{y}, f(\mathbf{x}; \mathbf{W}, \mathbf{A}))$, namely,
\begin{equation}
    \mathbf{A}_{t+\Delta} = \mathbf{A}_{t} - \eta \nabla\mathcal{H}(\mathbf{A}_{t}),
    \label{eqn:sgd}
\end{equation}
where $\eta$ is the learning rate and $\eta >> \eta_s$.
Then the proximal operator chooses a neighborhood point of $A_{t+\Delta}$ that minimizes the group sparsity regularization, \ie
\begin{align}
    \mathbf{A}_{t+1} & = \mathbf{prox}_{\lambda\eta\mathcal{R}}(\mathbf{A}_{t+\Delta}) \\ \nonumber
    &= \argmin_{\mathbf{A}} \left\{\mathcal{R}(\mathbf{A}_{t+\Delta}) + \frac{1}{2\lambda\eta} \|\mathbf{A}-  \mathbf{A}_{t+\Delta}\|_2^2 \right\}.
\end{align}

The sparsity regularizer $\Phi(\cdot)$ can have different forms, \eg, $\ell_1$ norm~\cite{parikh2014proximal}, $\ell_{1/2}$ norm~\cite{xu2012lregularization}, $\ell_{1-2}$ norm~\cite{yao2016fast}, and $\mathrm{logsum}$~\cite{gu2017weighted}. All of them try to approximate the $\ell_0$ norm. In this paper, we mainly use $\{\ell_{p}: {p=1, 1/2}\}$ regularizers while we also include the $\ell_{1-2}$ and $\mathrm{logsum}$ regularizers in the ablation studies.
The proximal operators of the four regularizers have a closed-form solution. Briefly, the solution is the soft-thresholding operator~\cite{boyd2011admm} for $p=1$ and the half-thresholding operator for $p=1/2$~\cite{xu2012lregularization}. The solutions are appended in the supplementary material. The gradient step and proximal step are interleaved in the optimization phase of the regularized loss until some predefined stopping criterion is achieved. After each epoch, groups with $\ell_2$ norms smaller than a predefined threshold $\mathcal{T}$ are nullified. And the compression ratio in terms of FLOPs is calculated. When the difference between the current and the target compression ratio $\gamma_c$ and $\gamma^\ast$ is lower than the stopping criterion $\alpha$, the compression phase stops.
The detailed compression algorithm that utilizes the proximal gradient is shown in Algorithm~\ref{alg:proximal_gradient}.

\subsection{Binary search of the nullifying threshold}
After the compression phase stops, the resulting compression ratio is not exactly the same as the target compression ratio. 
To fit the target compression ratio, we use a binary search algorithm to determine the nullifying threshold $\mathcal{T}$. 
The compression ratio $\gamma$ is actually a monotonous function of the threshold $\mathcal{T}$, \ie $\gamma = g(\mathcal{T})$. 
However, the explicit expression of the function $g(\cdot)$ is not known. 
Given a target compression threshold $\gamma^\ast$, we want to derive the threshold used to nullifying the sparse groups, \ie $\mathcal{T}^\ast = g^{-1}(\gamma^\ast)$, where $g^{-1}(\cdot)$ is the inverse function of $g(\cdot)$. 
The binary search approach shown in Algorithm~\ref{alg:binary_search} starts with an initial threshold $\mathcal{T}_0$ and a step $s$. 
It adjusts the threshold $\mathcal{T}$ according to the values of the current and target compression ratio. 
The step $s$ is halved if the target compression ratio sits between the previous one $\gamma_{n-1}$ and the current one $\gamma_n$. 
The searching procedure stops if final compression ratio $\gamma_n$ is closed enough to the target, 
\ie, $|\gamma_n - \gamma^\ast| \leq \mathcal{C}$.

\subsection{Gradient based adjustment of learning rate}

In the ResNet basic block, both of the two $3 \times 3$ convolutional layers are attached a sparsity-inducing matrix $\mathbf{A}^1$ and $\mathbf{A}^2$, namely, $1 \times 1$ convolutional layers. We empirically find that the gradient of the first sparsity-inducing matrix is larger than that of the second. Thus, it is easier for the first matrix to jump to a point with larger average group $\ell_2$ norms. This results in unbalanced compression of the two sparsity-inducing matrices since the same nullifying threshold is used for all of the layers. That is, much more channels of $\mathbf{A}^2$ are compressed than that of the first one. This is not a desired compression approach since both of the two layers are equally important. A balanced compression between them is preferred. 

To solve this problem, we adjust the learning rate of the first and second sparsity-inducing matrices according to their gradients. Let the ratio of the average group $\ell_2$ norm between the gradients of the matrices be
\begin{equation}
    \rho = \sum_{g}\left(\nabla\mathbf{A}^1\right)_g / \sum_{g}\left(\nabla \mathbf{A}^2\right)_g.
    \label{eqn:gradient_based_adjust}
\end{equation}
Then the learning rate of the first convolution is divided by $\rho ^ m$. We empirically set $m=1.35$.

\subsection{Group $\ell_2$ norm based layer balancing}
The proximal gradient method depends highly on the group sparsity term. That is, if the initial $\ell_2$ norm of a group is small, then it is highly likely that this group will be nullified. The problem is that the distribution of the group $\ell_2$ norm across different layers varies a lot, which can result in quite unbalanced compression of the layers. In this case, a quite narrow bottleneck could appear in the compressed network and would hamper the performance. To solve this problem, we use the mean of the group $\ell_2$ norm of a layer to recalibrate the regularization factor of the layer. That is,
\begin{equation}
    \lambda ^ l = \lambda  \frac{1}{G}\sum_{g=1}^{G} \|\mathbf{A}_g\|_2,
    \label{eqn:layer_balancing}
\end{equation}
where $\lambda^l$ is the regularization factor of the $l$-th layer. In this way, the layers with larger average group $\ell_2$ norm get a larger punishment.  

\subsection{Regularization factor annealing}

The compression procedure starts with a fixed regularization factor. However, towards the end of the  compression phase, the fixed regularization factor may be so large that more than the desired groups are nullified in an epoch. Thus, to solve the problem, we anneal the regularization factor when the average group $\ell_2$ norm shrinks below some threshold. The annealing also impacts the proximal step but has less influence while the gradient step plays a more active role in finding the local minimum.

\subsection{Distillation loss in the finetuning phase}
\label{subsec:distillation_loss}

In Eqn.~\ref{eqn:group_sparsity}, the prediction loss and the groups sparsity regularization are used to solve the compression problem. After the compression phase, the derived model is further finetuned. During this phase, a distillation loss is exploited to force similar logit outputs of the original network and the pruned one. The vanilla distillation loss is used, \ie
\begin{equation}\label{eqn:distillation_loss}
    \begin{split}
        \mathcal{L} &= (1 - \alpha)\mathcal{L}_{ce}(\mathbf{y}, \sigma(\mathbf{z}_c))\\
        &\quad +  2\alpha T ^2 \mathcal{L}_{ce}\left(\sigma\left(\frac{\mathbf{z}_c}{T}\right), \sigma\left(\frac{\mathbf{z}_o}{T}\right)\right),
    \end{split}
\end{equation}
where $\mathcal{L}_{ce}(\cdot)$ denotes the cross-entropy loss, $\sigma(\cdot)$ is the softmax function, $\mathbf{z}_c$ and $\mathbf{z}_o$ are the logit outputs of the compressed and the original network. For the sake of simplicity, the network parameters are omitted. We use a fixed balancing factor $\alpha = 0.4$ and temperature $T = 4$. 

\section{Implementation Considerations}
\label{sec:implementation}

\subsection{Sparsity-inducing matrix in network blocks}
In the analysis of Sec.~\ref{sec:method}, a $1 \times 1$ convolutional layer with the sparsity-inducing matrix is appended after the uncompressed layer. 
When it comes to different network blocks, we tweak it a little bit.
As stated, both of the $3 \times 3$ convolutions in the ResNet~\cite{he2016deep} basic block are appended with a $1 \times 1$ convolution. For the first sparsity-inducing matrix, group sparsity regularization can be enforced on either the columns or the rows of the matrix. As for the second matrix, group sparsity can be enforced on its rows due to the existence of the skip connection.

The ResNet~\cite{he2016deep} and ResNeXt~\cite{xie2017aggregated} bottleneck block has the structure of $1 \times 1 \rightarrow 3 \times 3 \rightarrow 1 \times 1$ convolutions. Here, the natural choice of sparsity-inducing matrices  are the leading and the ending convolutions. For the ResNet bottleneck block, the two matrices select the input and output channels of the middle $3 \times 3$ convolution, respectively. Things become a little bit different for the ResNeXt bottleneck since the middle $3 \times 3$ convolution is a group convolution. So the aim becomes enforcing sparsity on the already existing groups of the group convolution. In order to do that, the parameters related to the groups in the two sparsity-inducing matrices are concatenated. Then group sparsity is enforced on the new matrix. After the compression phase, a whole group can be nullified.  

\subsection{Initialization of $\mathbf{W}$ and $\mathbf{A}$}
For the ResNet and ResNeXt bottleneck block, $1 \times 1$ convolutions are already there. So the original network parameters are used directly. However, it is necessary to initialize the newly added sparsity-inducing matrix $\mathbf{A}$. Two initialization methods are tried. The first one initializes $\mathbf{W}$ and $\mathbf{A}$ with the pretrained parameters and identity matrix, respectively. The second method first calculates the singular value decomposition of $\mathbf{W}$, \ie $\mathbf{W} = \mathbf{U}\mathbf{S}\mathbf{V}^T$. Then the left eigenvector $\mathbf{U}$ and the matrix $\mathbf{S}\mathbf{V}^T$ are used to initialize $\mathbf{W}$ and $\mathbf{A}$. Note that the singular values are annexed by the right eigenvector. Thus, the columns of $\mathbf{W}$, \ie the filters of the convolutional layer ly on the surface of the unit sphere in the high-dimensional space.

\section{Experimental Results}
\label{sec:experimental_results}
In this section, the proposed method is validated on three image classification datasets including CIFAR10, CIFAR100~\cite{krizhevsky2009learning}, and ImageNet2012~\cite{deng2009imagenet}. 
The network compression method is applied to ResNet~\cite{he2016deep}, ResNeXt~\cite{xie2017aggregated}, VGG~\cite{simonyan2014very}, and DenseNet~\cite{huang2017densely} on CIFAR10 and CIFAR100, WRN~\cite{zagoruyko2016wide} on CIFAR100, and ResNet50 on ImageNet2012. 
For ResNet20 and ResNet56 on CIFAR dataset, the residual block is the basic ResBlock with two $3 \times 3$ convolutional layers.
For ResNet164 on CIFAR and ResNet50 on ImageNet, the residual block is a bottleneck block.
The investigated models of ResNeXt are ResNeXt20 and ResNeXt164 with carlinality 32, and bottleneck width 1. WRN has 16 convolutional layers with widening factor 10.

The training protocol of the original network is as follows.
The networks are trained for 300 epochs with SGD on CIFAR dataset. 
The momentum is $0.9$ and the weight decay factor is $10^{-4}$. 
Batch size is $64$. The learning rate starts with $0.1$ and decays by $10$ at Epoch 150 and 225. The ResNet50 model is loaded from the pretrained PyTorch model~\cite{paszke2017automatic}.
The models are trained with Nvidia Titan Xp GPUs.
The proposed network compression method is implemented by PyTorch.
We fix the hyper parameters of the proposed method by empirical studies. 
The stop criterion $\alpha$ in Algorithm~\ref{alg:proximal_gradient} is set to 0.1. The threshold $\mathcal{T}$ is set to 0.005. Unless otherwise stated, $\ell_1$ regularizer is used and the regularization factor is set to $2e^{-4}$. As already mentioned, during the compression step, we set different learning rates for $\mathbf{W}$ and $\mathbf{A}$. The ratio between $\eta_s$ and $\eta$ is 0.01.

\begin{figure}
\begin{minipage}[c]{0.5\linewidth}
  \vspace*{\fill}
  \centering
  \includegraphics[width=1.0\linewidth]{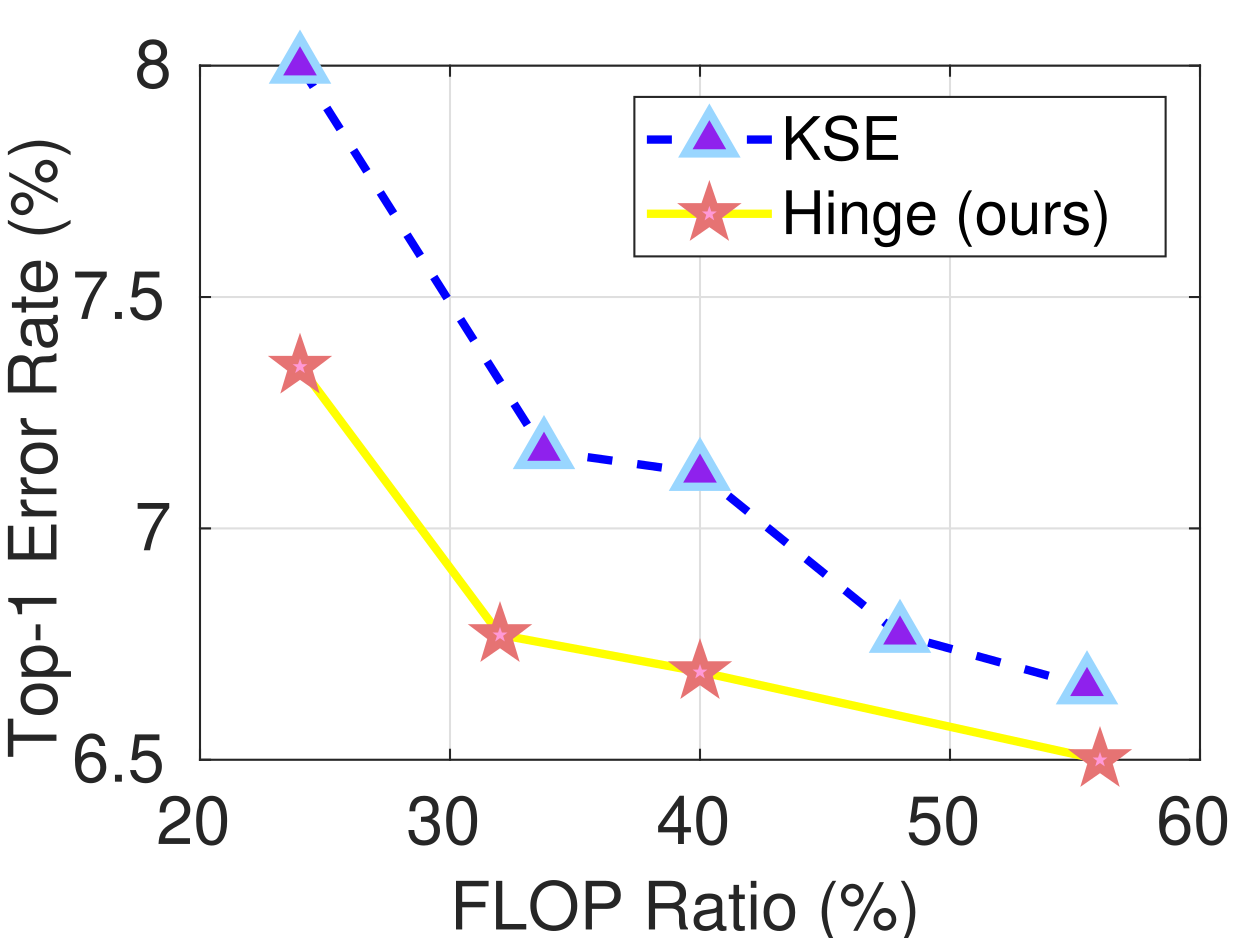}
  \subcaption{FLOP ratio comparison.}
  \label{fig:hinge_kse_flop}
\end{minipage}%
\begin{minipage}[c]{0.5\linewidth}
  \vspace*{\fill}
  \centering
  \includegraphics[width=1.0\linewidth]{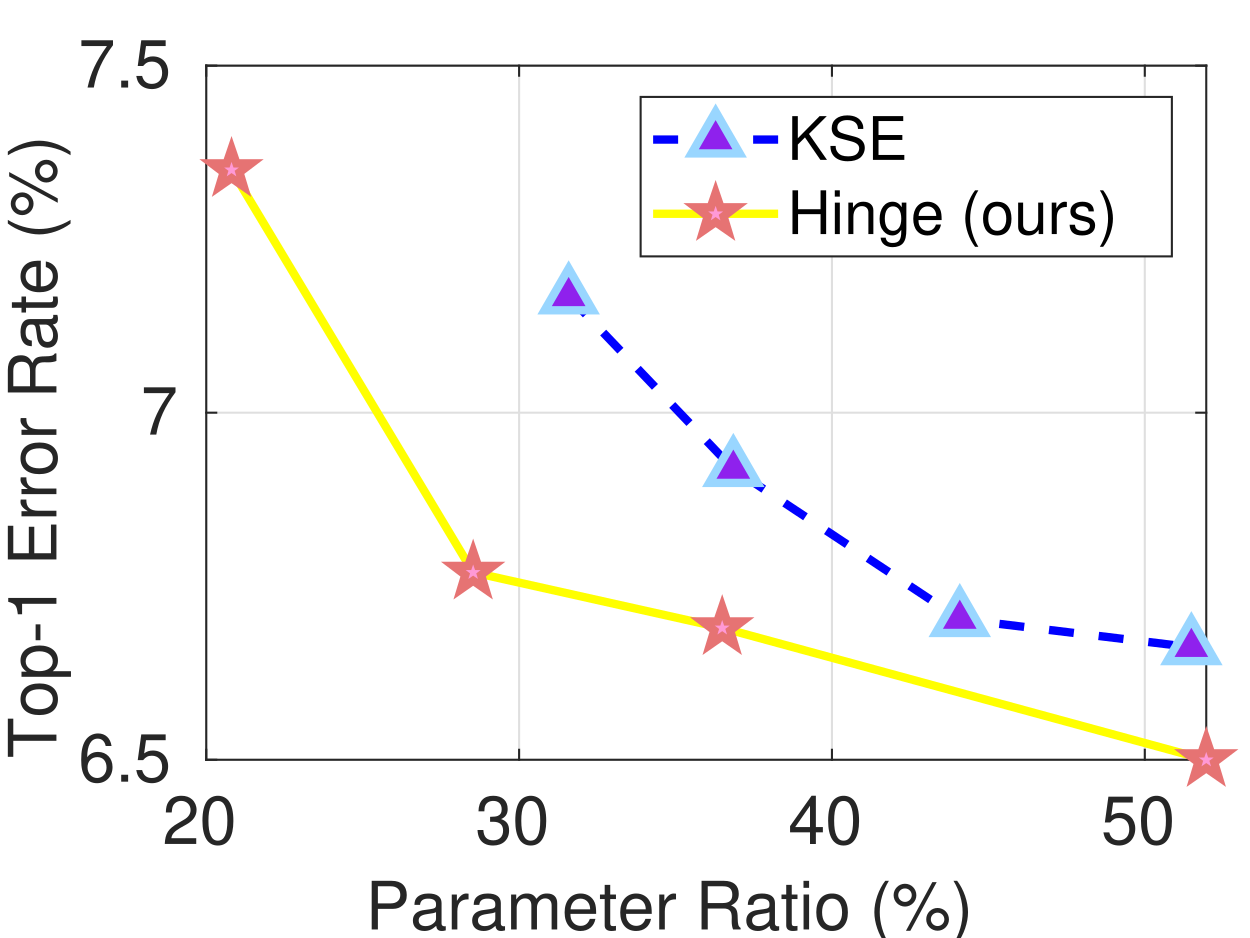}
  \subcaption{Parameter ratio comparison.}
  \label{fig:hinge_kse_param}
\end{minipage}%
\vspace{-0.2cm}
\caption{(a) FLOP and (b) parameter comparison between KSE~\cite{Li_2019_CVPR_KSE} and Hinge under different compression ratio. ResNet56 is compressed. Top-1 error rate is reported.}
\label{fig:hinge_kse}
\vspace{-0.4cm}
\end{figure}

\begin{figure}
\begin{minipage}[c]{0.5\linewidth}
  \vspace*{\fill}
  \centering
  \includegraphics[width=1.0\linewidth]{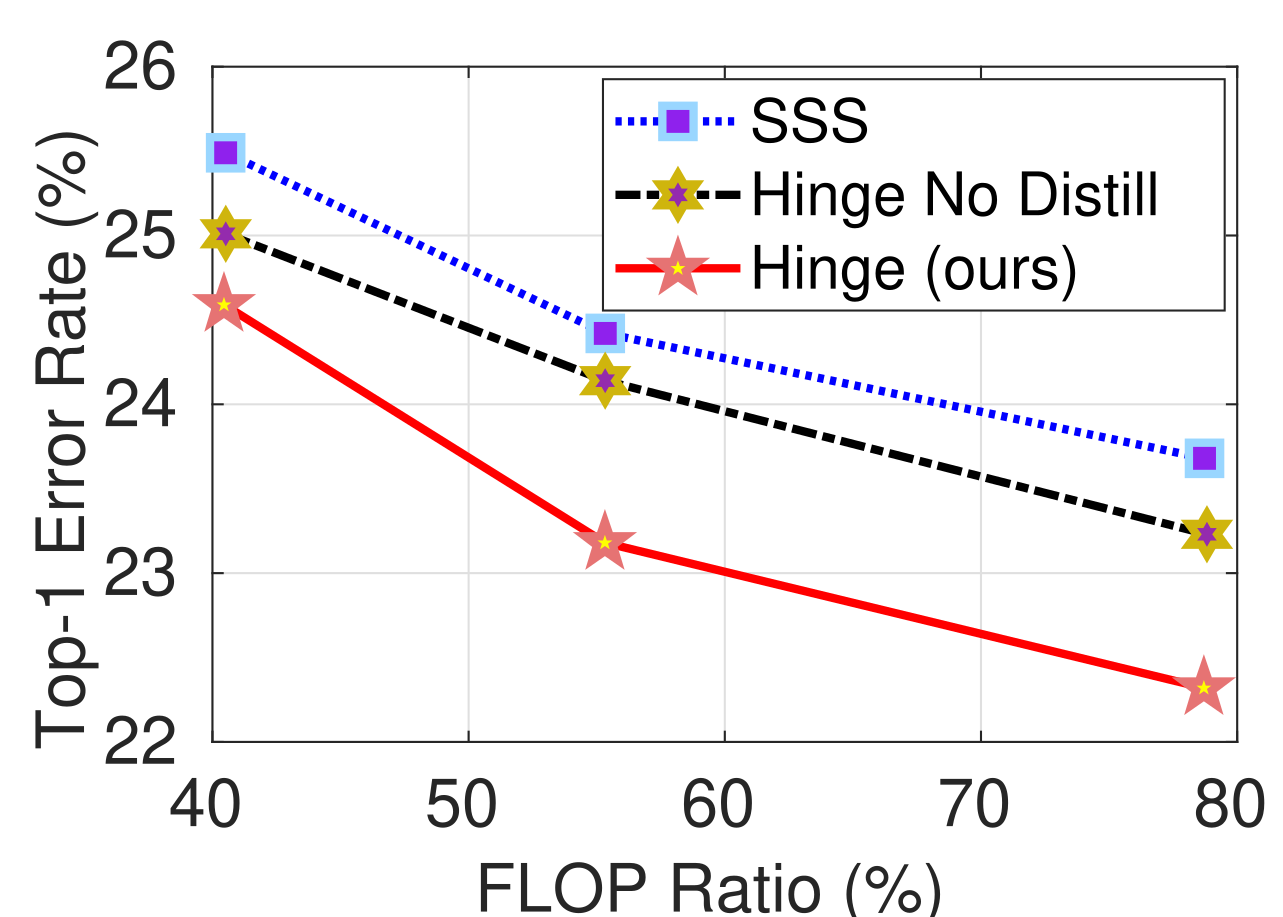}
  \subcaption{ResNet164}
  \label{fig:resnet164_cifar100}
\end{minipage}%
\begin{minipage}[c]{0.5\linewidth}
  \vspace*{\fill}
  \centering
  \includegraphics[width=1.0\linewidth]{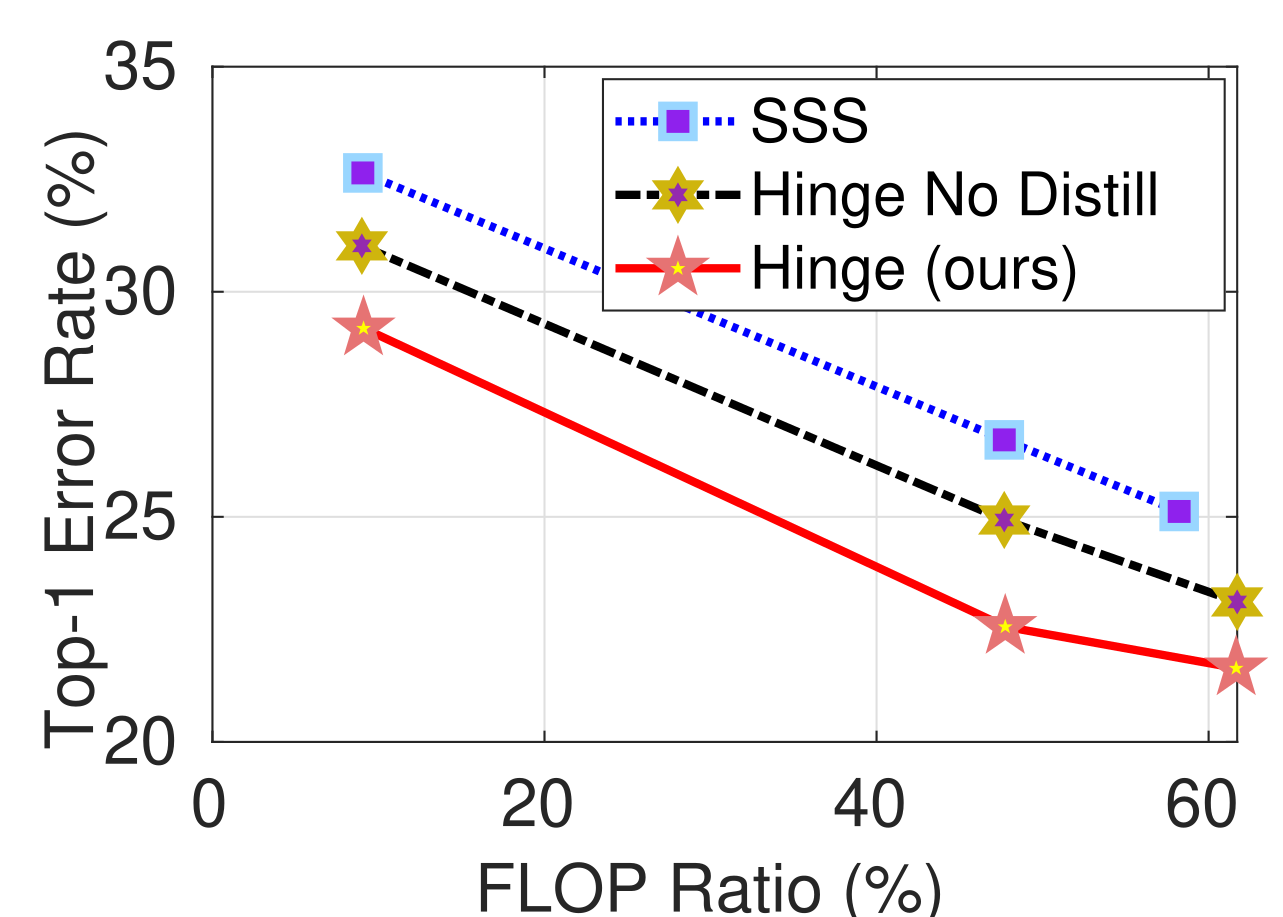}
  \subcaption{ResNeXt164}
  \label{fig:resnext164_cifar100}
\end{minipage}%
\vspace{-0.2cm}
\caption{Comparison between SSS~\cite{huang2018data} and the proposed method. Top-1 error rate is reported for CIFAR100.}
\label{fig:layer164_cifar100}
\vspace{-0.4cm}
\end{figure}

\begin{figure*}
\begin{minipage}[c]{.25\textwidth}
  \vspace*{\fill}
  \centering
  \includegraphics[width=1.04\linewidth]{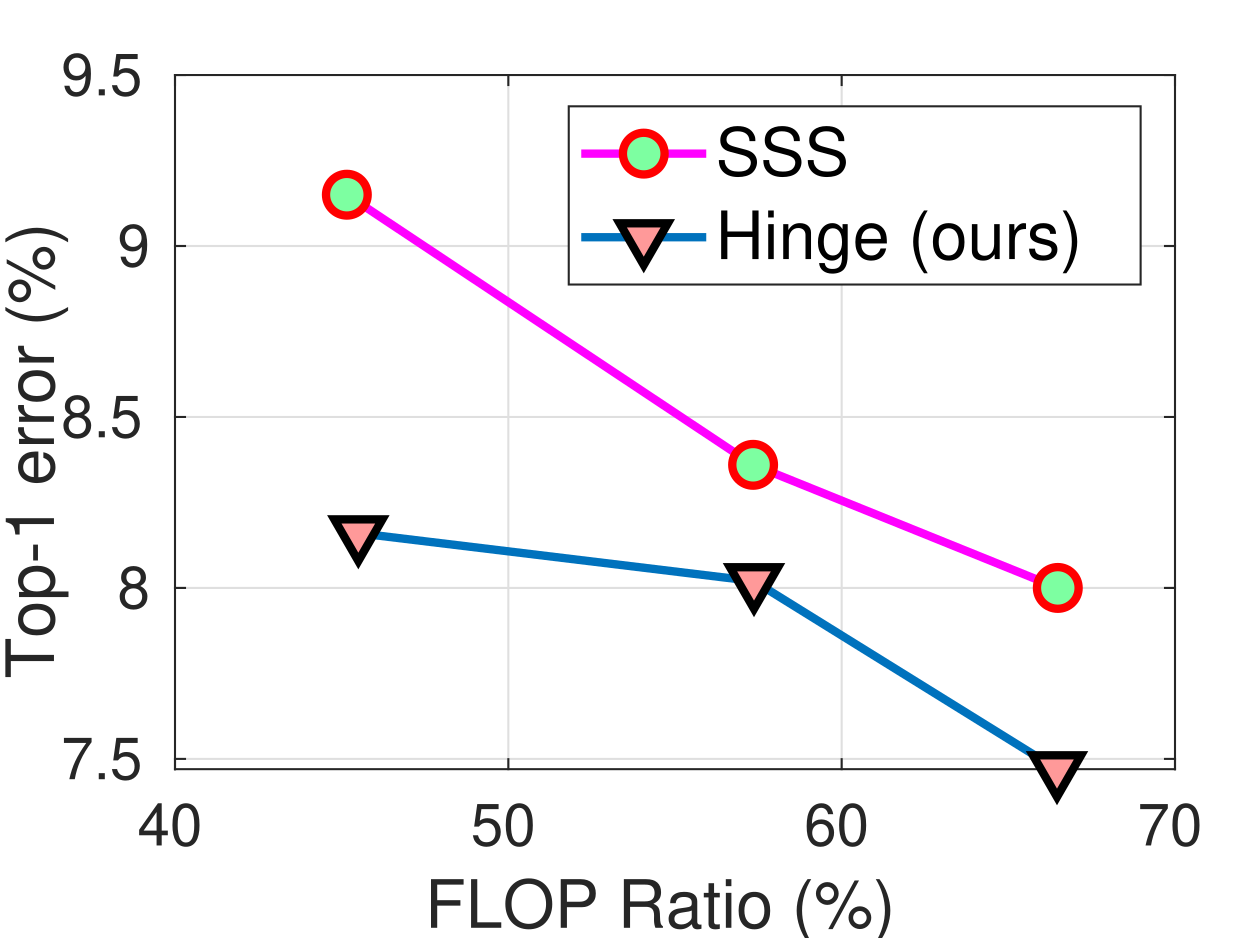}
  \subcaption{CIFAR10, ResNet20}
  \vspace{-0.2cm}
  \label{fig:cifar10_resnet20}
\end{minipage}%
\begin{minipage}[c]{.25\textwidth}
  \vspace*{\fill}
  \centering
  \includegraphics[width=1.04\linewidth]{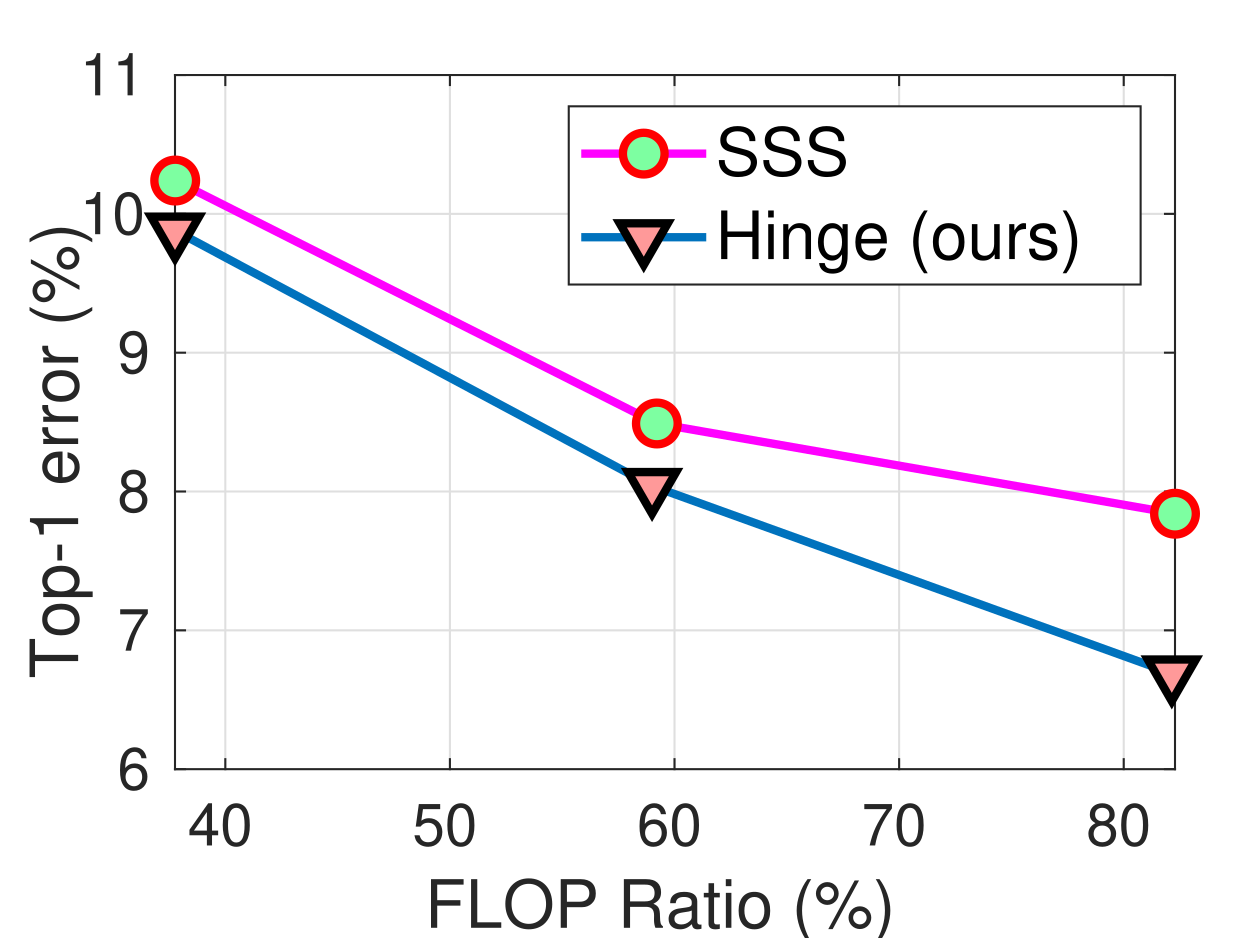}
  \subcaption{CIFAR10, ResNeXt20}
  \vspace{-0.2cm}
  \label{fig:cifar10_resnext20}
\end{minipage}%
\begin{minipage}[c]{.25\textwidth}
  \vspace*{\fill}
  \centering
  \includegraphics[width=1.04\linewidth]{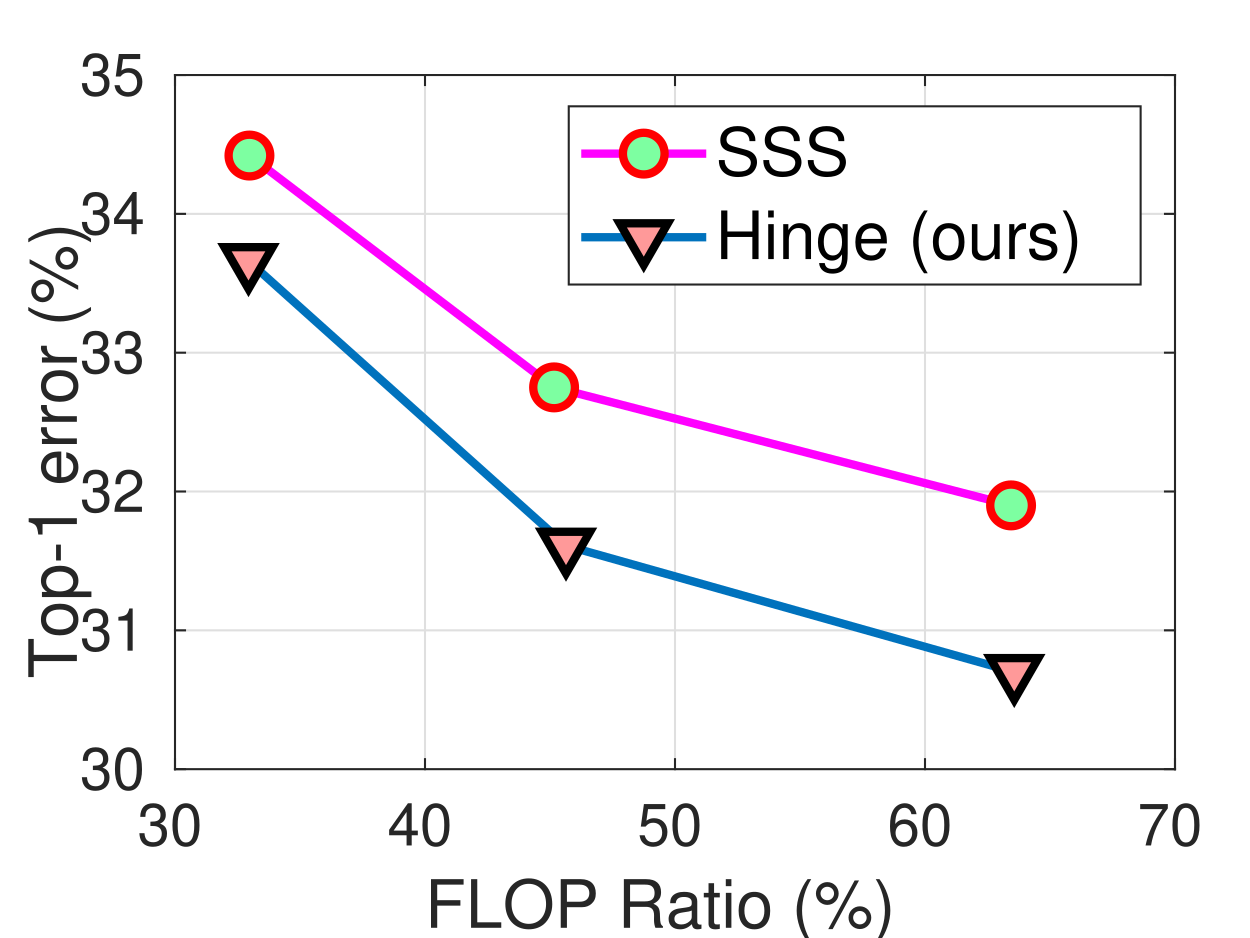}
  \subcaption{CIFAR100, ResNet20}
  \vspace{-0.2cm}
  \label{fig:cifar100_resnet20}
\end{minipage}%
\begin{minipage}[c]{.25\textwidth}
  \vspace*{\fill}
  \centering
  \includegraphics[width=1.04\linewidth]{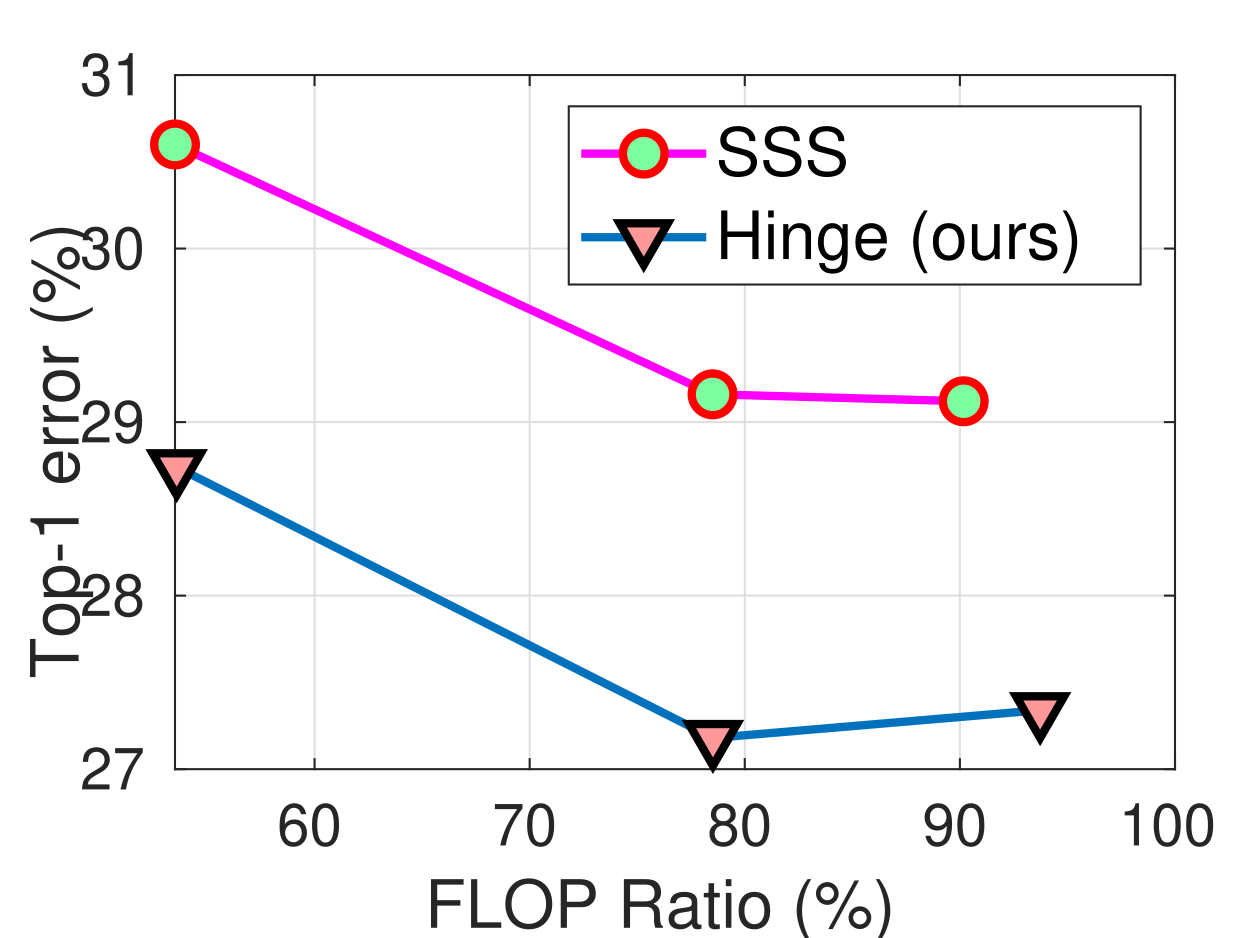}
  \subcaption{CIFAR100, ResNeXt20}
  \vspace{-0.2cm}
  \label{fig:cifar100_resnext20}
\end{minipage}%
\caption{Comparison between SSS~\cite{huang2018data} and the proposed method. Top-1 error rate is reported. (a) and (b) shows the results on CIFAR10 while (c) and (d) shows the results on CIFAR100.}
    \vspace{-0.4cm}
\label{fig:sss_hinge_comparison}
\end{figure*}

\begin{table}[t]
    \scriptsize
    \begin{center}
        \begin{tabular}{c|c|c|c|c}
        \toprule
        Model & Method & Top-1 / BL (\%)  & FLOPs (\%) & Params (\%)\\ \midrule
        \multirow{11}{*}{\shortstack{ResNet- \\56}} & \cite{zhao2019variational}& 7.74/6.96 & 79.70 & 79.51 \\
            & GAL-0.6~\cite{lin2019towards}         & 6.62/7.64 & 63.40 & 88.20 \\
            & \cite{li2017pruning}      & 6.94/6.96 & 62.40 & 86.30 \\
            & NISP~\cite{yu2018nisp}                & 6.99/6.96 & 56.39 & 57.40 \\
            & CaP~\cite{Minnehan_2019_CVPR_CaP} &6.78 / 6.49 & 50.20 & --\\
        & ENC~\cite{Kim_2019_CVPR_ENC} &7.00 / 6.90 & 50.00 & --\\
        & AMC~\cite{He_2018_ECCV_AMC} &8.10 / 7.20 & 50.00 & -- \\
        & KSE~\cite{Li_2019_CVPR_KSE} &6.77 / 6.97 & 48.00 & 45.27\\
        & FPGM~\cite{he2019filter} &6.74 / 6.41 & 47.70 & -- \\
        & Hinge (ours) & \textbf{6.31} / 7.05 & 50.00 & 48.73 \\ \cline{2-5}
        & KSE~\cite{Li_2019_CVPR_KSE} & 8.00 / 6.97  & 24.00 & -- \\
        & Hinge (ours) & \textbf{7.35} / 7.05 & 24.00 & 20.80\\ \midrule
        \multirow{3}{*}{\shortstack{ResNet- \\20}} &  \cite{zhao2019variational} &8.34 / 7.99 & 83.53 &79.59 \\ 
        & SSS~\cite{huang2018data} & 9.15 / 7.47 & 45.16 &83.41 \\
        & Hinge (ours) & \textbf{8.16} / 7.46 & 45.50 &44.55\\ \midrule
        \multirow{2}{*}{\shortstack{ResNet- \\ 164}} & SSS~\cite{huang2018data} & 5.78 / 5.18 & 53.53 &84.75\\
        & Hinge (ours) & \textbf{5.4} / 4.97 & 53.61 &70.34\\ \midrule
        \multirow{2}{*}{\shortstack{ResNeXt- \\20}} & SSS~\cite{huang2018data} & 8.49 / 7.08 & 59.21 &76.57\\
        & Hinge (ours) & \textbf{8.04} / 7.46 & 59.00 &63.95\\ \midrule
        \multirow{2}{*}{\shortstack{ResNeXt- \\ 164}} & SSS~\cite{huang2018data} & 5.42 / 6.41 & 44.38&64.38\\
        & Hinge (ours) & \textbf{5.13} / 4.82 & 44.42 &50.53\\ \midrule
        \multirow{3}{*}{VGG16} & \cite{zhao2019variational} & 6.82 / 6.75 & 60.90&26.66\\
        & GAL-0.1~\cite{lin2019towards} & 6.58 / 6.04 & 54.80 & 17.80 \\
        & Hinge (ours) & \textbf{6.41} / 5.98 & 60.93 & 19.95 \\ \midrule
        \multirow{3}{*}{\shortstack{DenseNet- \\ 12-40}} & GAL-0.01~\cite{lin2019towards} & 5.39 / 5.19 & 64.70 & 64.40 \\
        & \cite{zhao2019variational} & 6.84 / 5.89 & 55.22&40.33\\
        & Hinge (ours) & \textbf{5.33} / 5.26 & 55.60 & 72.46\\
        \bottomrule
        \end{tabular}
    \end{center}
    \vspace{-0.3cm}
    \caption{Comparison of CIFAR10 compression results. ``FLOPs'' and ``Params'' denote the remaining percentage of FLOP and parameter quantities of the compressed models and the lower the better. The other tables and figures follows the same convention.}
    \label{tbl:cifar10}
    \vspace{-0.15cm}
\end{table}

\begin{table}[t]
    \scriptsize
    \begin{center}
        \begin{tabular}{c|c|c|c|c}
        \toprule
        Model & Method & Top-1 / BL (\%)  & FLOPs (\%) & Params (\%)\\ \midrule
        \multirow{6}{*}{WRN} & CGES~\cite{yoon2017combined} & 21.97 / 21.62 & 75.56 & -- \\
        & Hinge-NA & 23.61 / 21.58 & 75.59 & 84.31\\
        & Hinge (ours) & \textbf{21.79} / 21.58 & 75.61 & 83.29\\ \cline{2-5}
        & CGES~\cite{yoon2017combined} & 22.75 / 21.62 & 57.31 & -- \\
        & Hinge-NA & 23.13 / 21.58 & 57.41 & 68.72\\
        & Hinge (ours) & \textbf{22.06} / 21.58 & 57.39 & 67.80\\ \midrule
        \multirow{2}{*}{ResNet20} & SSS~\cite{huang2018data} & 34.42 / 30.91 & 32.98 &54.42\\
        & Hinge (ours) & \textbf{33.66} / 31.17 & 32.94 &33.64\\ \midrule
        \multirow{2}{*}{ResNet164} & SSS~\cite{huang2018data} & 24.42 / 23.31 & 55.33 &86.75\\
        & Hinge (ours) & 23.12 / 23.22 & 55.32&76.57\\ \midrule
        \multirow{2}{*}{ResNeXt20} & SSS~\cite{huang2018data} & 30.60 / 28.00 & 53.51 &76.34\\
        & Hinge (ours) & \textbf{28.74} / 28.05 & 53.59 &65.24\\ \midrule
        \multirow{2}{*}{ResNeXt164} & SSS~\cite{huang2018data} & 26.71 / 23.18 & 47.69 &72.47\\
        & Hinge (ours) & \textbf{22.56} / 23.13 & 47.75 &58.49\\
        \bottomrule
        \end{tabular}
    \end{center}
    \vspace{-0.3cm}
    \caption{Comparison of CIFAR100 compression results. For a fair comparison, the model size from different methods is kept to the same level. Hinge-NA stands for our hinge method without regularization factor annealing during the compression phase.}
    \label{tbl:cifar100}
    \vspace{-0.4cm}
\end{table}

\subsection{Results on CIFAR10}

The experimental results on CIFAR10 are shown in Table.~\ref{tbl:cifar10}. The Top-1 error rate, the percentage of the remaining FLOPs and parameters of the compressed models are listed in the table. For ResNet56, two operating points are reported. The operating point of 50\% FLOP compression is investigated by a bunch of state-of-the-art compression methods. Our proposed method achieves the best performance under this constraint. At the compression ratio of 24\%, our approach is clearly better than KSE~\cite{Li_2019_CVPR_KSE}. 
For ResNet and ResNeXt with 20 and 164 layers, our method shoots a lower error rate than SSS. 
For VGG and DenseNet, the proposed method reduces the Top-1 error rate by 0.41 \% and 1.51 \% compared with \cite{zhao2019variational}.
In Fig.~\ref{fig:hinge_kse}, we compare the FLOPs and number of parameters of the compressed model by KSE and the proposed method under different compression ratios. As shown in the figure, our compression method outperforms KSE easily.
Fig.~\ref{fig:cifar10_resnet20} and \ref{fig:cifar10_resnext20} show more comparison between SSS and our methods on CIFAR10. Our method forms a lower bound for SSS.

The ablation study on ResNet56 is shown in Table~\ref{tbl:reset56_abalation}. Different combinations of the hyper parameters $\mathcal{T}$ and $\alpha$ are investigated. There are only slight changes in the results for different combinations. Anyway, when $\mathcal{T} = 0.005$ and $\alpha = 0.01$, our method achieves the lowest error rate. And we use this combination for the other experiments. As for the different regularizers, $\ell_1$ and $\ell_{1/2}$ regularization are clearly better than $\ell_{1-2}$ and $\mathrm{logsum}$. Due to the simplicity of the solution to the proximal operator of $\ell_1$ regularization, we use $\ell_1$ instead of $\ell_{1/2}$ in the other experiments. 

\subsection{Results on CIFAR100}

Table~\ref{tbl:cifar100} shows the compression results on CIFAR100. 
For the compression of WRN, we analyze the influence of regularization factor annealing during the compression phase. It is clear that with the annealing mechanism, the proposed method achieves much better performance. This is because towards the end of the compression phase, the proximal gradient solver has found quite a good neighbor of the local minimum. In this case, the regularization factor should diminish in order for a better exploration around the local minimum. Compared with the previous group sparsity method CGES~\cite{yoon2017combined}, our hinge method with the annealing mechanism results in better performance.

Fig.~\ref{fig:layer164_cifar100} compares the SSS and our method for the 164-layer networks. Even without the distillation loss, our method is already better than SSS. When the distillation loss is utilized, the proposed method brings the Top-1 error rate to an even lower level. The corresponding results for the 20-layer networks are shown in Fig.~\ref{fig:cifar100_resnet20} and \ref{fig:cifar100_resnext20}, respectively.

\begin{table}[t]
    \small
    \begin{center}
        \begin{tabular}{c|c|c|c}
        \toprule
        Regularizer & Threshold $\mathcal{T}$ & $\alpha$ & Top-1 error (\%) \\ \midrule
        $\ell_1$ & 0.001 & 0.05 & 6.54 \\
        $\ell_1$ & 0.005 & 0.1  & 6.53 \\
        $\ell_1$ & 0.001 & 0.05 & 6.66 \\
        $\ell_1$ & 0.005 & 0.01 & 6.37 \\ \midrule
        $\mathrm{logsum}$ & 0.005 & 0.01 & 6.53 \\
        $\ell_{1/2}$ & 0.005 & 0.01 & 6.31 \\
        $\ell_{1-2}$ & 0.005 & 0.01 & 6.56 \\
        \bottomrule
        \end{tabular}
    \end{center}
    \vspace{-0.15cm}
    \caption{Ablation study: the proposed compression method is applied to ResNet56 and tested on CIFAR10. The compression ratio is fixed to $50\%$. Different regularizers and hyper parameters $\mathcal{T}$ and $\alpha$ are examined.}
    \label{tbl:reset56_abalation}
    \vspace{-0.4cm}
\end{table}

\begin{table}[t]
    \small
    \begin{center}
        \begin{tabular}{c|c|c}
        \toprule
        Method & Top-1 Error  & FLOPs (\%) \\ \midrule
        
        SSS~\cite{huang2018data} & 25.82  & 68.55\\
        ThinNet-70~\cite{luo2017thinet} & 27.96 & 63.21 \\
        NISP~\cite{yu2018nisp} & 28.01 & 55.99 \\
        Taylor-56\%~\cite{molchanov2019importance} & 25.50 & 55.01 \\
        FPGM~\cite{he2019filter} & 25.17  & 47.50\\
        \textbf{Hinge (ours)} & 25.30 &46.55  \\
        RRBP~\cite{zhou2019accelerate} &27.00 & 45.45 \\
        GAL~\cite{lin2019towards} & 28.20 & 44.98 \\
        \bottomrule
        \end{tabular}
    \end{center}
    \vspace{-0.15cm}
    \caption{Results of compressing ResNet50 on ImageNet2012. Entries are sorted according to FLOPs.}
    \label{tbl:resnet50_imagenet}
    \vspace{-0.4cm}
\end{table}

\subsection{Results on ImageNet}

The comparison results of compressing ResNet50 on the ImageNet2012 dataset is shown in Table~\ref{tbl:resnet50_imagenet}. Since different methods compare the compressed models under different FLOP compression rates, it is only possible to compare different methods under roughly comparable compression rates. Compared with those methods, our method achieves state-of-the-art trade-off performance between Top-1 error rate and FLOP compression ratio.   
\section{Conclusion}
\label{sec:conclusion}
In this paper, we propose to hinge filter pruning and decomposition via group sparsity. By enforcing group sparsity regularization on the different structured groups, \ie, columns and rows of the sparsity-inducing matrix, the manipulation of the tensor breaks down to filter pruning and decomposition, respectively. The unified formulation enables the devised algorithm to flexibly switch between the two modes of network compression, depending on the specific circumstances in the network. Proximal gradient method with gradient based learning rate adjustment, layer balancing, and regularization factor annealing are used to solve the optimization problem. Distillation loss is used in the finetuning phase. The experimental results validate the proposed method. 

\noindent\textbf{Acknowledgments:}
This work was partly supported by the ETH Z\"urich Fund (OK), by VSS ASTRA, SBB and Huawei projects, and by Amazon AWS and Nvidia GPU grants.

{\small
\bibliographystyle{ieee_fullname}
\bibliography{egbib}
}

\appendix
\onecolumn
\clearpage


\maketitle
{\centering
\section*{Supplementary Material for ``Group Sparsity: The Hinge Between Filter Pruning and Decomposition for Network Compression''}}

\section{Closed-form Solutions to the Proximal Operators}
\label{eqn:closed_form_solutions}
The proximal operator of a given function $f(\cdot)$ is defined by 
\begin{equation}
    \mathbf{prox}_{\lambda f} = \argmin_v \left\{f(v) - \frac{1}{\lambda} \|x -v \|_2^2 \right\}
    \label{eqn:proximal_operator}
\end{equation}
This operator has closed-form solution when the function $f(\cdot)$ has the form of $\ell_1$, $\ell_{1/2}$, $\ell_{1-2}$, and $\mathrm{logsum}$ regularization. For $\ell_1$, the solution is the soft-thresholding function and for $\ell_{1/2}$ it is the so-called half-thresholding function. The soft-thresholding function is defined as
\begin{equation}
    \mathcal{S}_{\lambda}(x) = \mathrm{sgn}(x)[|x| - \lambda]_+,
    \label{eqn:soft_thresholding}
\end{equation}
where $\mathrm{sgn}(\cdot)$ is the sign function and $[\cdot]_+$ calculates the maximum of the argument and 0. The hard-thresholding function is given by
\begin{equation}
    \mathcal{H}_\lambda(x) = 
    \begin{cases} 
    \frac{2}{3}x(1 + \cos(\frac{2\pi}{3} - \frac{2}{3}\phi_\lambda(x))), & |x| > \frac{\sqrt[3]{54}}{4}(\lambda)^\frac{2}{3},
    \\ 
    0, & \text{otherwise},
    \end{cases}
    \label{eqn:half_thresholding}
\end{equation}
where $\phi_\lambda(x) = \arccos(\frac{\lambda}{8}(\frac{|x|}{3})^{-\frac{3}{2}})$.

The $\ell_{2,1}$ group sparsity regularizer is defined as 
\begin{equation}
    \mathcal{R}(\mathbf{A}) = \Phi(\|\mathbf{A}_g\|_2) =  {\sum_{g}\|\mathbf{A}_g \|_2^p},
    \label{eqn:lp_regularizer}
\end{equation}
where $\Phi(\cdot)$ is the function of the group $\ell_2$ norms $\|\mathbf{A}_g\|_2$ and has the form of $\ell_1$ norm here.
The proximal operator of the sparsity-inducing matrix $\mathbf{A}$ defined in the main paper is
\begin{equation}
        \mathbf{A}_{t+1}  = \mathbf{prox}_{\lambda\eta\mathcal{R}}(\mathbf{A}_{t+\Delta}) = \argmin_{\mathbf{A}} \left\{\mathcal{R}(\mathbf{A}_{t+\Delta}) + \frac{1}{2\lambda\eta} \|\mathbf{A}-  \mathbf{A}_{t+\Delta}\|_F^2 \right\},
        \label{eqn:proximal_operator_group_sparsity}
\end{equation}
where the function $\mathcal{R}(\cdot)$ replaces $f(\cdot)$ in Eqn.~\ref{eqn:proximal_operator}.
The closed-form solution of the proximal operator in Eqn.~\ref{eqn:proximal_operator_group_sparsity} can be derived from the solutions to Eqn.~\ref{eqn:proximal_operator} according to the following theorem~\cite{beck2017first}.

\begin{theorem}
Let $f:\mathbb{E} \rightarrow \mathbb{R}$ be a function given by $f(\mathbf{x}) = g(\|\mathbf{x}\|)$, where $g:\mathbb{R} \rightarrow (-\infty, \infty]$ is a proper closed and convex function satisfying $\mathrm{dom}(g) \subseteq [0, \infty)$. Then,
\begin{equation}
    \mathbf{prox}_{\lambda f}(\mathbf{x}) = 
    \begin{cases} 
    \mathbf{prox}_{\lambda g}(\|\mathbf{x}\|_2)\frac{\mathbf{x}}{\|\mathbf{x}\|_2}, & \mathbf{x} \neq \mathbf{0},
    \\ 
    \{\mathbf{u} \in \mathbb{E}: \|\mathbf{u}\|_2 = \mathbf{prox}_{\lambda g}(\mathbf{0})\}, & \mathbf{x} = \mathbf{0}.
    \end{cases}
    \label{eqn:half_thresholding}
\end{equation}
\label{theorem:theorem1}
\end{theorem}

Thus, with a little bit variable substitution, when $\Phi(\cdot)$ is $\ell_1$ regularizer, the solution to Eqn.~\ref{eqn:proximal_operator_group_sparsity} is given by 
\begin{equation}
    \mathbf{A}_{t+1} = \left[1- \frac{\lambda \eta}{\|\mathbf{A}_g\|_2} \right]_{+}\mathbf{A}_{g,i},
    \label{eqn:solution_l1}
\end{equation}
where $\mathbf{A}_{g,i}$ is the $i$-th element in the $g$-th group of the sparsity-inducing matrix $\mathbf{A}$, and for the sake of simplicity, the subscript $_{t + \Delta}$ is omitted. 

When the function $\Phi(\cdot)$ has the form of $\ell_{1/2}$, $\ell_{1-2}$, and $\mathrm{logsum}$, it is non-convex. However, we still use the variable substitution in Theorem~\ref{theorem:theorem1} experimentally and the corresponding results in the main paper are also very competitive.
For $\ell_{1/2}$ regularizer, the solution is given by
\begin{equation}
    \mathbf{A}_{t+1} =
    \begin{cases} 
    \frac{2}{3}\left(1 + \cos\left(\frac{2\pi}{3} - \frac{2}{3}\phi_{\lambda \eta}\left(\|\mathbf{A}_g\|_2\right)\right)\right)\mathbf{A}_{g,i}, & \|\mathbf{A}_g\|_2 > \frac{\sqrt[3]{54}}{4}(\lambda \eta)^\frac{2}{3},
    \\ 
    0,& \text{otherwise},
    \end{cases}
    \label{eqn:solution_l0.5}
\end{equation}
where $\phi_{\lambda \eta}(\|\mathbf{A}_g\|_2) = \arccos(\frac{\lambda \eta}{8}(\frac{\|\mathbf{A}_g\|_2}{3})^{-\frac{3}{2}})$. Similarly, the solution to the $\mathrm{logsum}$ regularizer is given by
\begin{equation}
    \mathbf{A}_{t+1} =
    \begin{cases} \frac{c_1 + \sqrt{c_2}}{2} \frac{\mathbf{A}_{g,i}}{\|\mathbf{A}_g\|_2},& c_2 >0,
    \\ 
    0,& c_2 \leqslant 0,
    \end{cases}
    \label{eqn:solution_l0.5}
\end{equation}
where $\lambda > 0$, $0 < \epsilon < \sqrt{\lambda \eta}$, $c_1 = \|\mathbf{A}_g\|_2 - \epsilon$, and $c_2 = c_1^2 - 4(\lambda \eta - \epsilon \|\mathbf{A}_g\|_2)$. When the regularizer is $\ell_{1-2}$ regularizer, then the solution is given by
\begin{equation}
    \mathbf{A}_{t+1} = (1 + \frac{\lambda \eta}{\|\mathbf{c}\|_2})[1-\frac{\lambda \eta}{\|\mathbf{A}_g\|_2}]_+\mathbf{A}_{g,i}
    \label{eqn:solution_l0.5}
\end{equation}
where $\mathbf{c}_g = [\|\mathbf{A}_g\|_2 - \lambda \eta]_+$. Note that the case where all of the group $\ell_2$ norms $\mathbf{A}_g$ equal 0 is not considered~\cite{yao2016fast} because it never happens during the optimization of our algorithm. The solutions are summarized in Table~\ref{tbl:solution}.

\begin{table}[t]
    \begin{center}
        \begin{tabular}{c|l}
        \toprule
        Regularizer & Solution \\ \midrule
        $\ell_1$ & $\mathbf{A}_{t+1} = \left[1- \frac{\lambda \eta}{\|\mathbf{A}_g\|_2} \right]_{+}\mathbf{A}_{g,i}$ \\ \midrule
        \multirow{2}{*}{$\ell_{1/2}$} & $ \mathbf{A}_{t+1} =
                            \begin{cases} 
                            \frac{2}{3}\left(1 + \cos\left(\frac{2\pi}{3} - \frac{2}{3}\phi_{\lambda \eta}\left(\|\mathbf{A}_g\|_2\right)\right)\right)\mathbf{A}_{g,i}, & \|\mathbf{A}_g\|_2 > \frac{\sqrt[3]{54}}{4}(\lambda \eta)^\frac{2}{3},
                            \\ 
                            0,& \text{otherwise},
                            \end{cases}
                            $\\ 
        & $\phi_{\lambda \eta}(\|\mathbf{A}_g\|_2) = \arccos(\frac{\lambda \eta}{8}(\frac{\|\mathbf{A}_g\|_2}{3})^{-\frac{3}{2}})$ \\ \midrule
        \multirow{2}{*}{$\ell_{1-2}$} & $\mathbf{A}_{t+1} = (1 + \frac{\lambda \eta}{\|\mathbf{c}\|_2})[1-\frac{\lambda \eta}{\|\mathbf{A}_g\|_2}]_+\mathbf{A}_{g,i}$\\ 
        & $\mathbf{c}_g = [\|\mathbf{A}_g\|_2 - \lambda \eta]_+$ \\ \midrule
        \multirow{2}{*}{$\mathrm{logsum}$} & $\mathbf{A}_{t+1} =
                                \begin{cases} \frac{c_1 + \sqrt{c_2}}{2} \frac{\mathbf{A}_{g,i}}{\|\mathbf{A}_g\|_2},& c_2 >0,
                                \\ 
                                0,& c_2 \leqslant 0,
                                \end{cases}$\\
        & $\lambda > 0$, $0 < \epsilon < \sqrt{\lambda \eta}$, $c_1 = \|\mathbf{A}_g\|_2 - \epsilon$, $c_2 = c_1^2 - 4(\lambda \eta - \epsilon \|\mathbf{A}_g\|_2)$ \\\bottomrule
        \end{tabular}
    \end{center}
    \vspace{-0.15cm}
    \caption{The solution to the proximal operator for $\ell_1$, $\ell_{1-2}$, $\ell_{1/2}$, and $\mathrm{logsum}$ regularizers.}
    \label{tbl:solution}
\end{table}

\begin{table}[t]
    \begin{center}
        \begin{tabular}{c|c|c|c|c}
        \hline
        Regularizer & $\ell_1$ & $\ell_{1-2}$ & $\ell_{1/2}$ & $\mathrm{logsum}$ \\ \hline
        Regularization factor $\lambda$ & $2e^{-4}$ & $2e^{-4}$ & $4e^{-4}$ & $9e^{-5}$\\ \hline
        \end{tabular}
    \end{center}
    \vspace{-0.15cm}
    \caption{The regularization factor for $\ell_1$, $\ell_{1-2}$, $\ell_{1/2}$, and $\mathrm{logsum}$ regularizers.}
    \label{tbl:regularization_factor}
    \vspace{-0.4cm}
\end{table}

\begin{figure*}[t]
\begin{minipage}[c]{1\textwidth}
  \vspace*{\fill}
  \centering
  \includegraphics[width=0.9\linewidth]{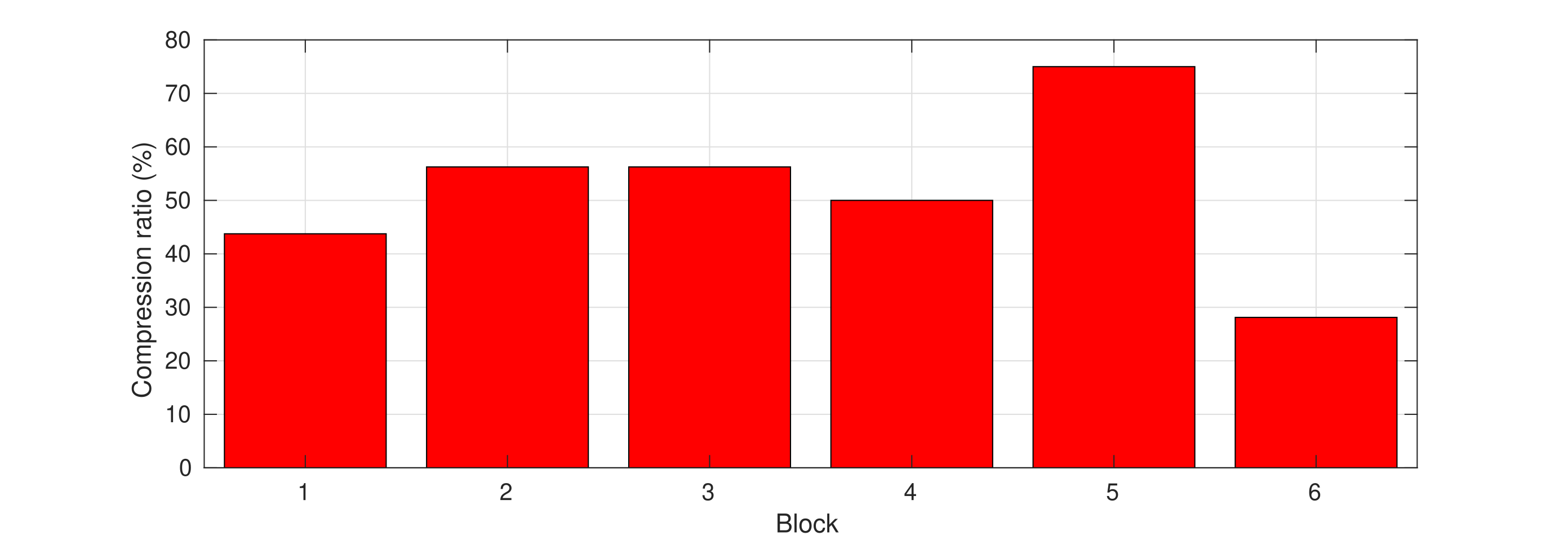}
  \subcaption{ResNeXt20~\cite{xie2017aggregated}, CIFAR100.}
  \vspace{-0.1cm}
  \label{fig:ratio_resnext20}
\end{minipage}%

\begin{minipage}[c]{1\textwidth}
  \vspace*{\fill}
  \centering
  \includegraphics[width=0.9\linewidth]{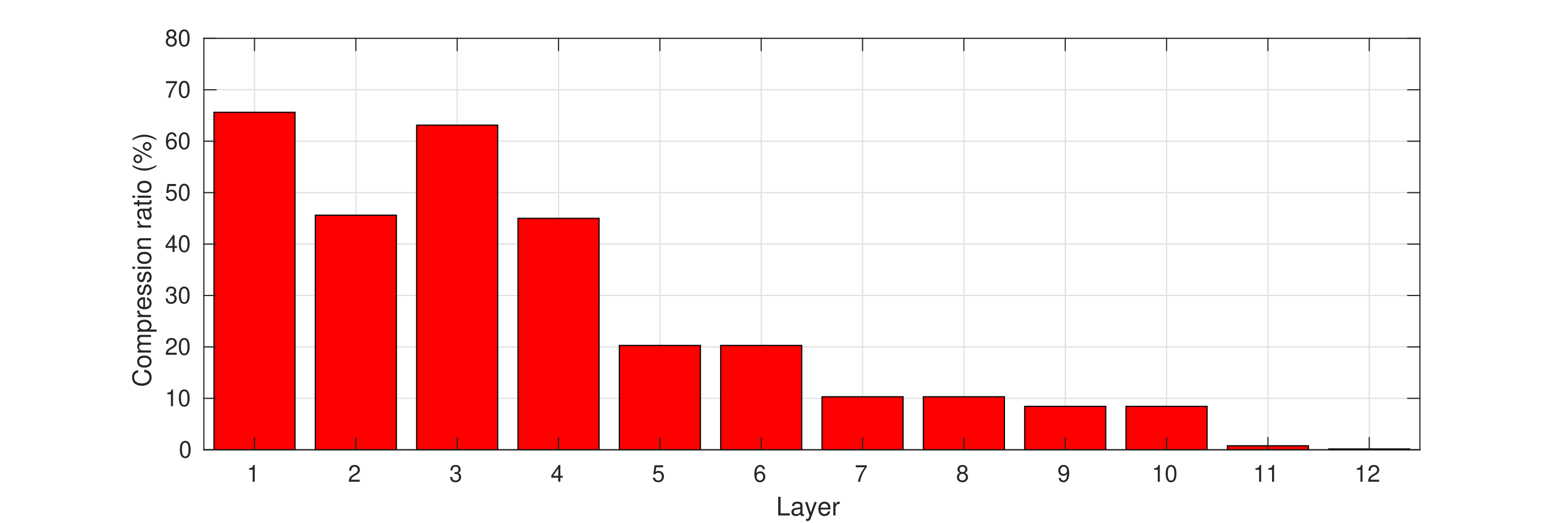}
  \subcaption{WRN~\cite{zagoruyko2016wide}, CIFAR100.}
  \vspace{-0.1cm}
  \label{fig:ratio_wrn}
\end{minipage}%

\begin{minipage}[c]{1\textwidth}
  \vspace*{\fill}
  \centering
  \includegraphics[width=0.9\linewidth]{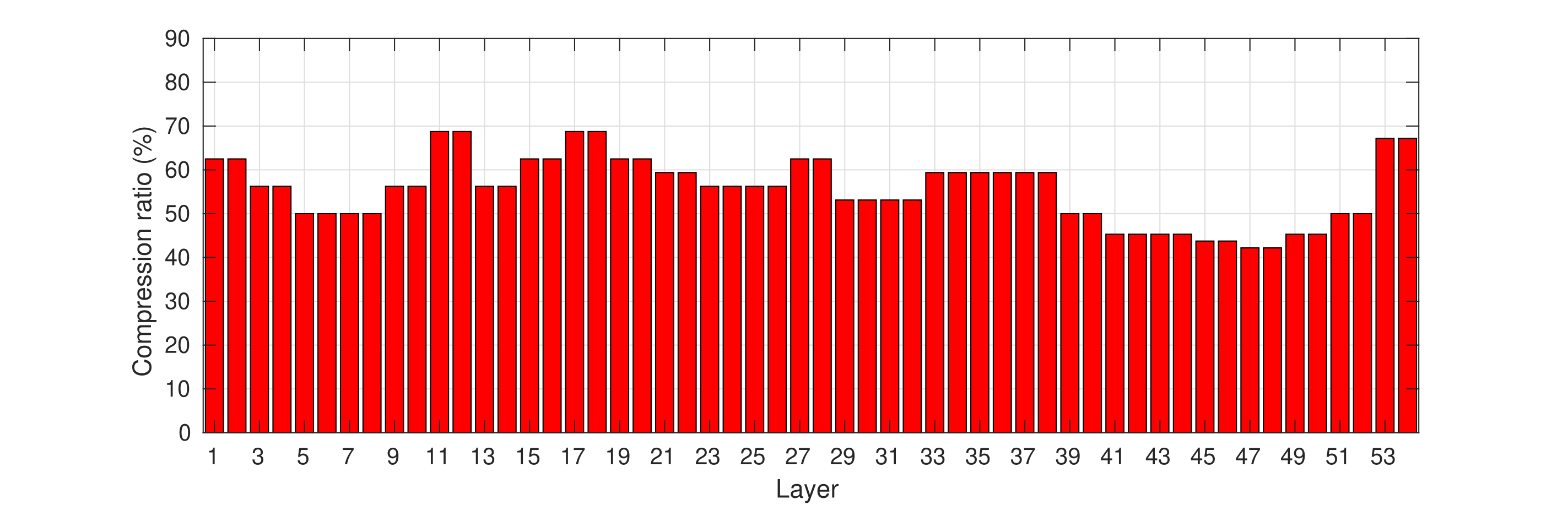}
  \subcaption{ResNet56~\cite{he2016deep}, CIFAR10.}
  \vspace{-0.2cm}
  \label{fig:ratio_resnet56}
\end{minipage}%

\caption{The layer-wise or block-wise compression ratio of the model resulting from the proposed method.}
\vspace{-0.4cm}
\label{fig:ratio1}
\end{figure*}

\begin{figure*}[t]
\begin{minipage}[c]{1\textwidth}
  \vspace*{\fill}
  \centering
  \includegraphics[width=0.9\linewidth]{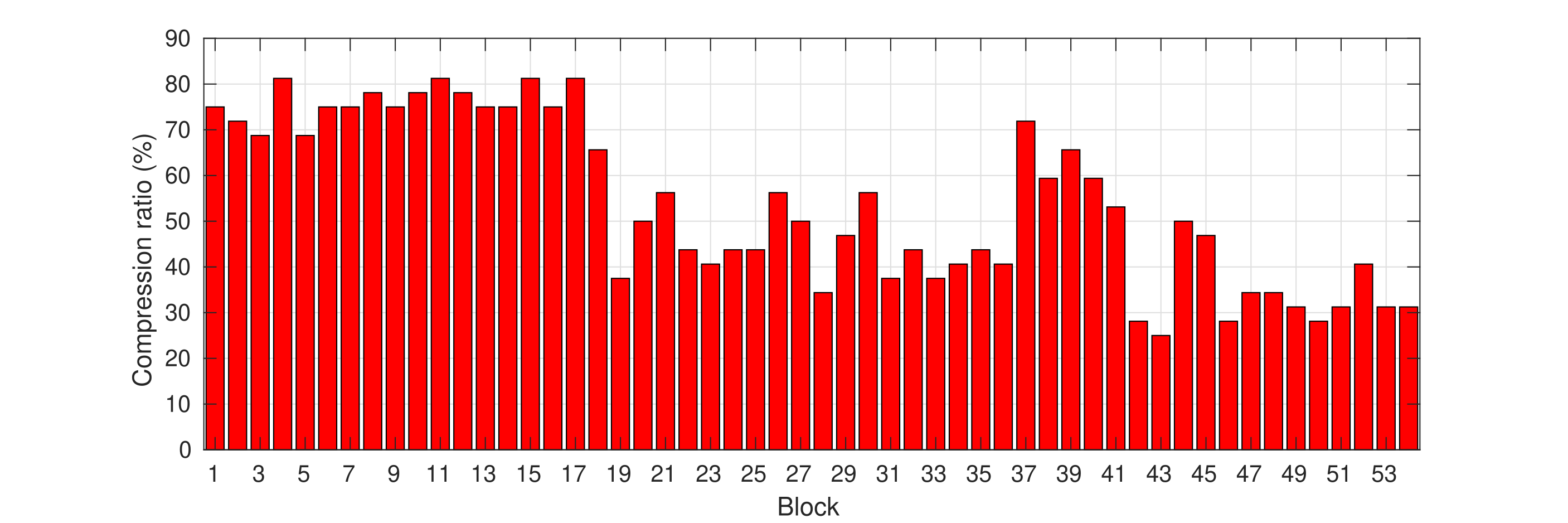}
  \subcaption{ResNeXt164~\cite{xie2017aggregated}.}
  \vspace{-0.1cm}
  \label{fig:ratio_resnext164}
\end{minipage}%

\begin{minipage}[c]{1\textwidth}
  \vspace*{\fill}
  \centering
  \includegraphics[width=0.9\linewidth]{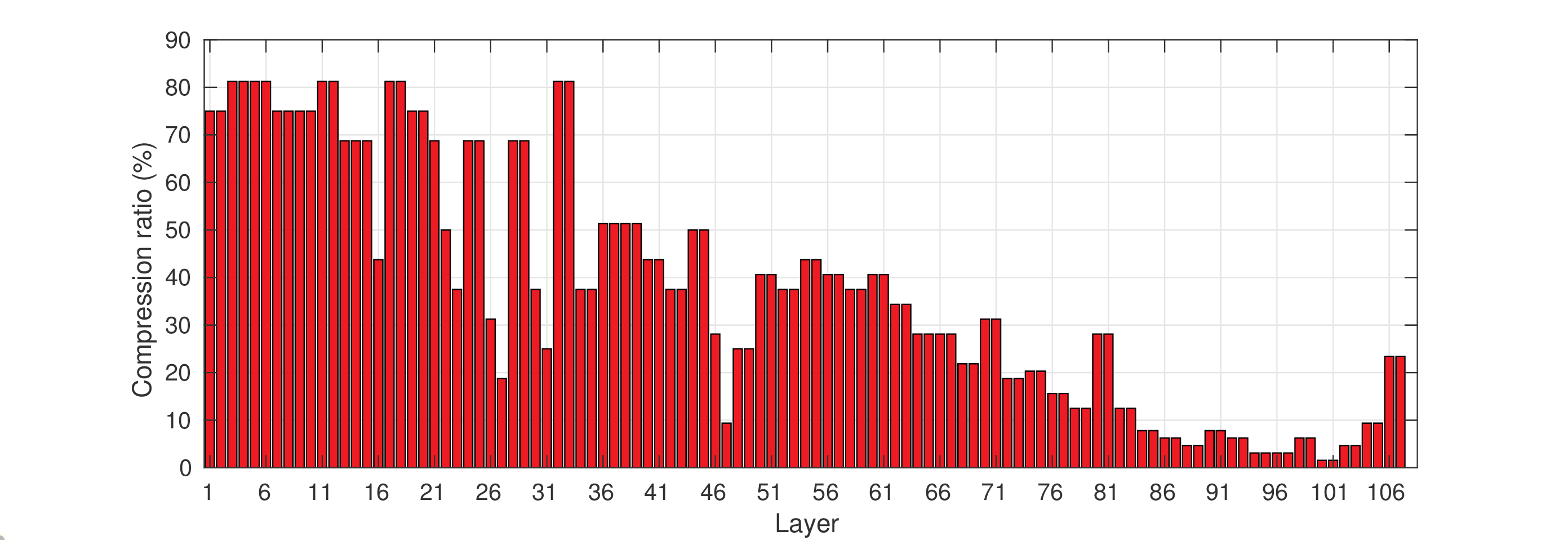}
  \subcaption{ResNet164~\cite{he2016deep}.}
  \vspace{-0.2cm}
  \label{fig:ratio_resnet164}
\end{minipage}%
\caption{The layer-wise or block-wise compression ratio of the model resulting from the proposed method. All results are reported for CIFAR100.}
\vspace{-0.4cm}
\label{fig:ratio2}
\end{figure*}

\begin{figure}
\begin{minipage}[c]{0.5\linewidth}
  \vspace*{\fill}
  \centering
  \includegraphics[width=0.5\linewidth]{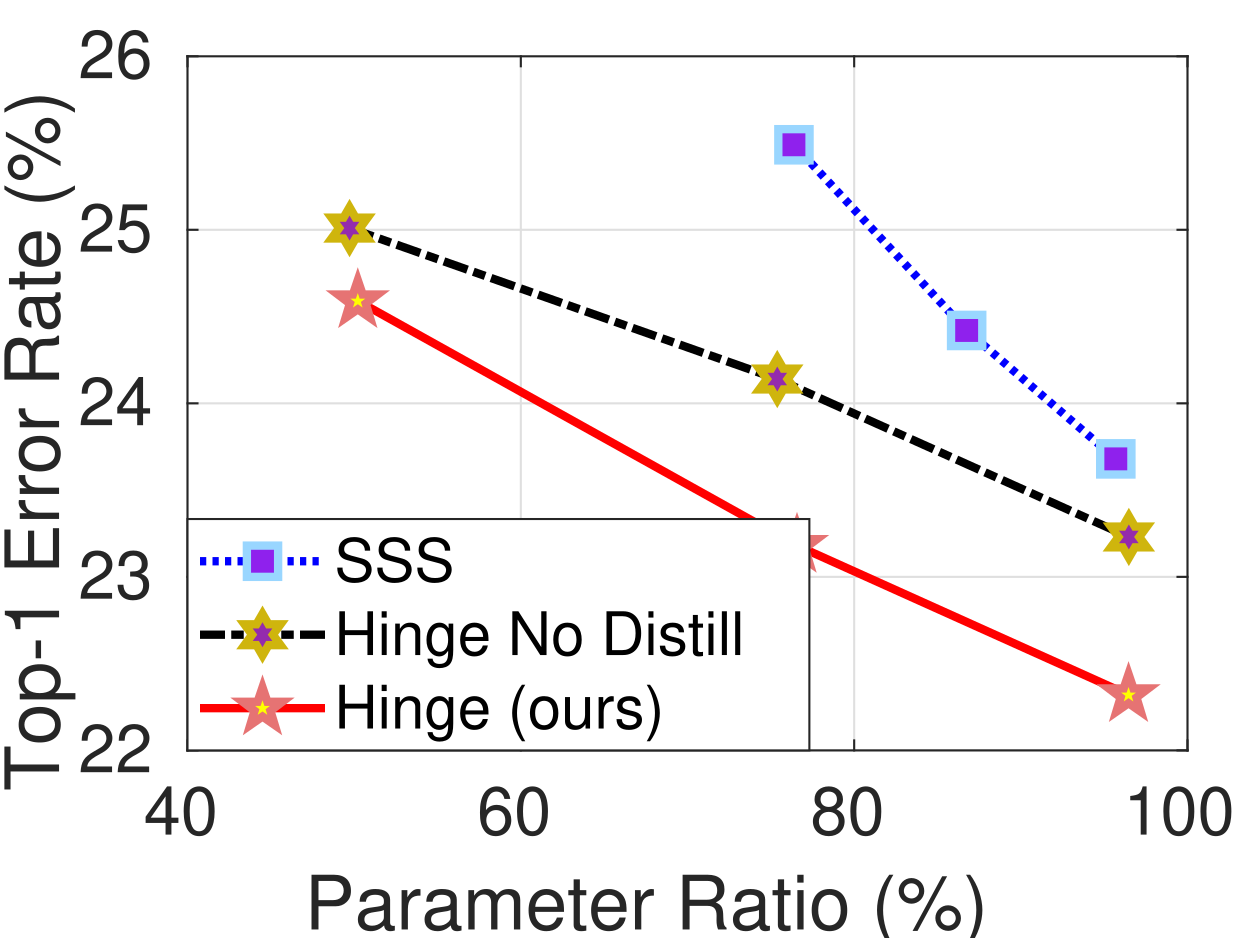}
  \subcaption{ResNet164}
  \label{fig:resnet164_cifar100}
\end{minipage}%
\begin{minipage}[c]{0.5\linewidth}
  \vspace*{\fill}
  \centering
  \includegraphics[width=0.5\linewidth]{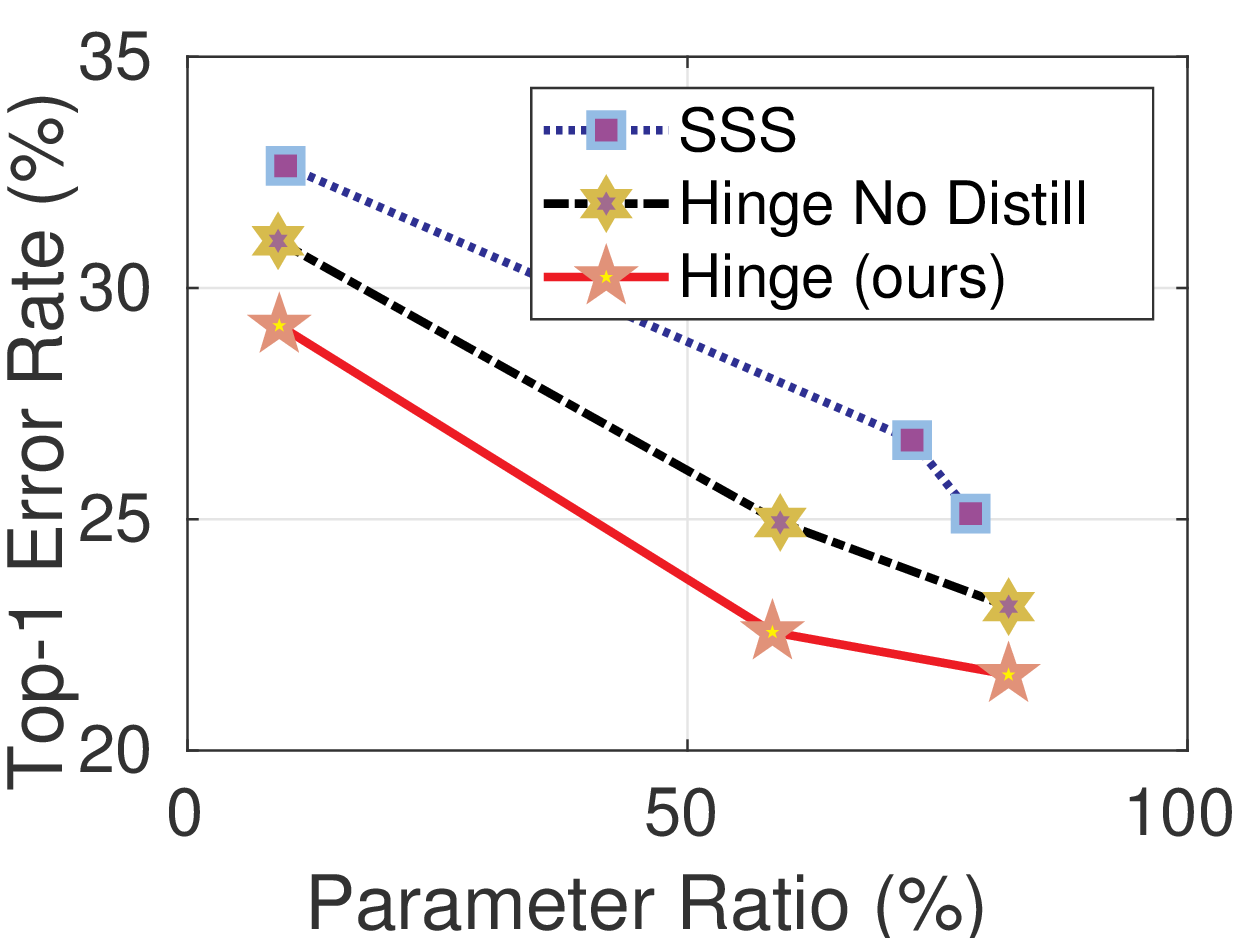}
  \subcaption{ResNeXt164}
  \label{fig:resnext164_cifar100}
\end{minipage}%
\vspace{-0.2cm}
\caption{Comparison between SSS~\cite{huang2018data} and the proposed method. Top-1 error rate is reported for CIFAR100.}
\label{fig:layer164_cifar100_params}
\vspace{-0.4cm}
\end{figure}

\begin{figure*}
\begin{minipage}[c]{.25\textwidth}
  \vspace*{\fill}
  \centering
  \includegraphics[width=1.04\linewidth]{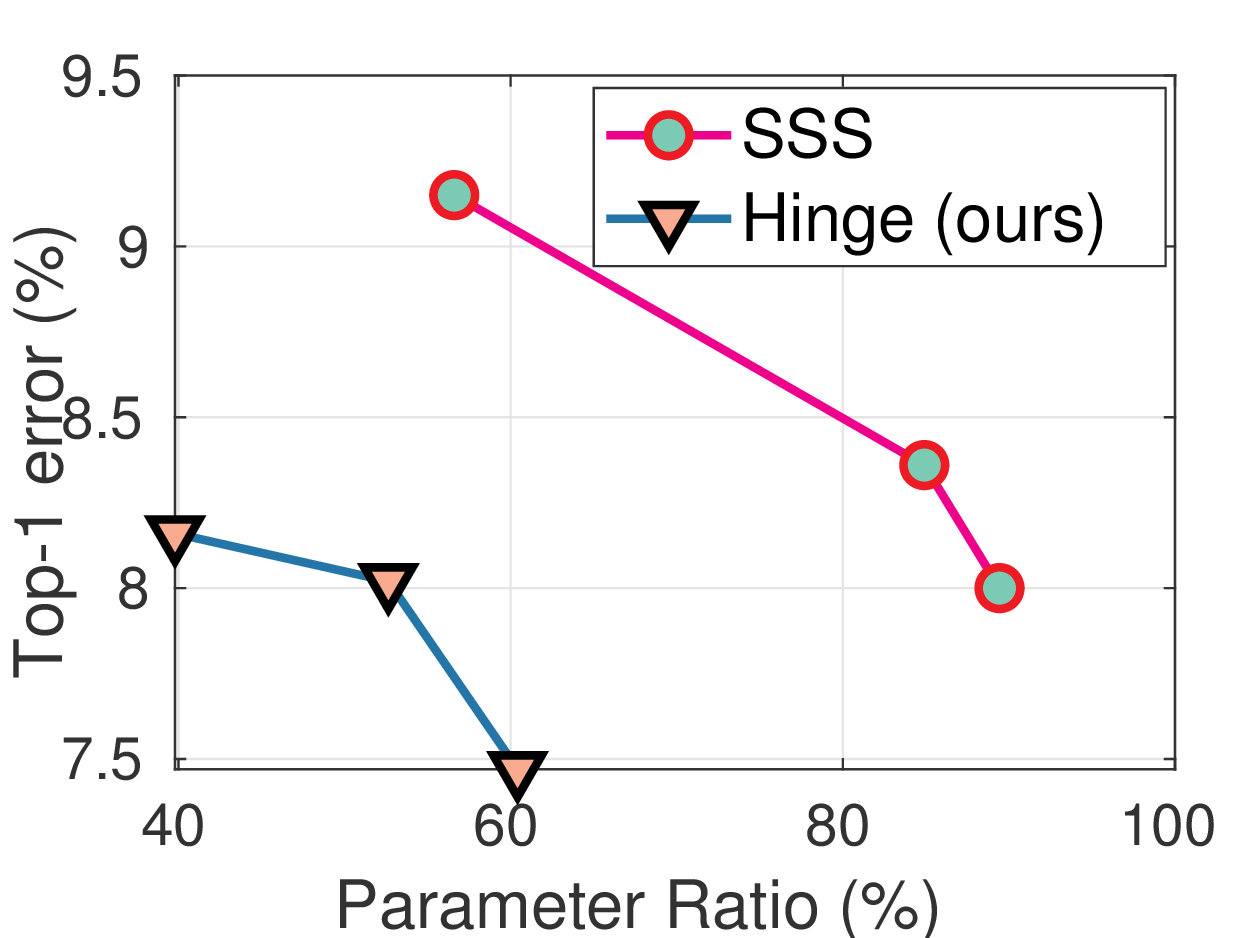}
  \subcaption{CIFAR10, ResNet20}
  \vspace{-0.2cm}
  \label{fig:cifar10_resnet20}
\end{minipage}%
\begin{minipage}[c]{.25\textwidth}
  \vspace*{\fill}
  \centering
  \includegraphics[width=1.04\linewidth]{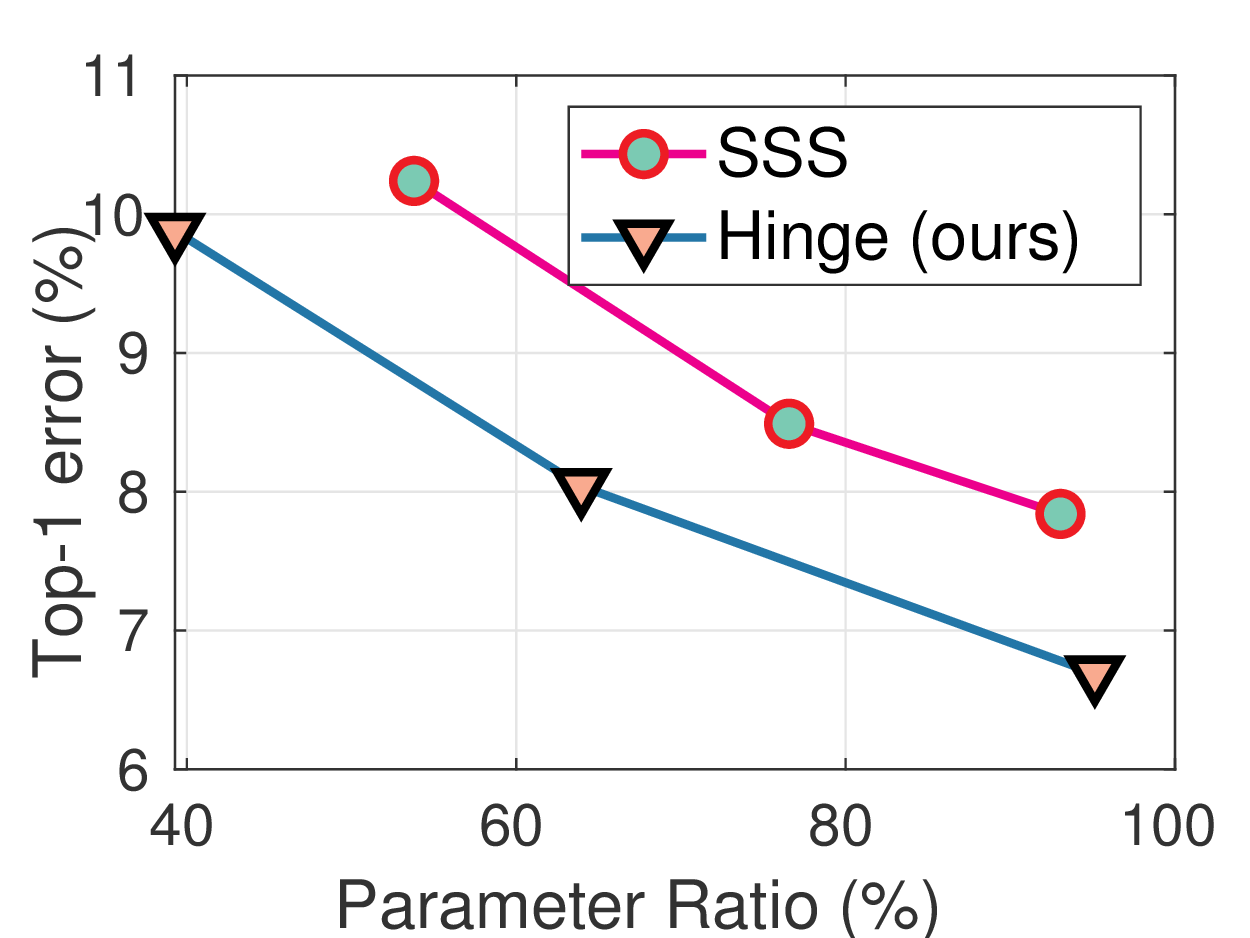}
  \subcaption{CIFAR10, ResNeXt20}
  \vspace{-0.2cm}
  \label{fig:cifar10_resnext20}
\end{minipage}%
\begin{minipage}[c]{.25\textwidth}
  \vspace*{\fill}
  \centering
  \includegraphics[width=1.04\linewidth]{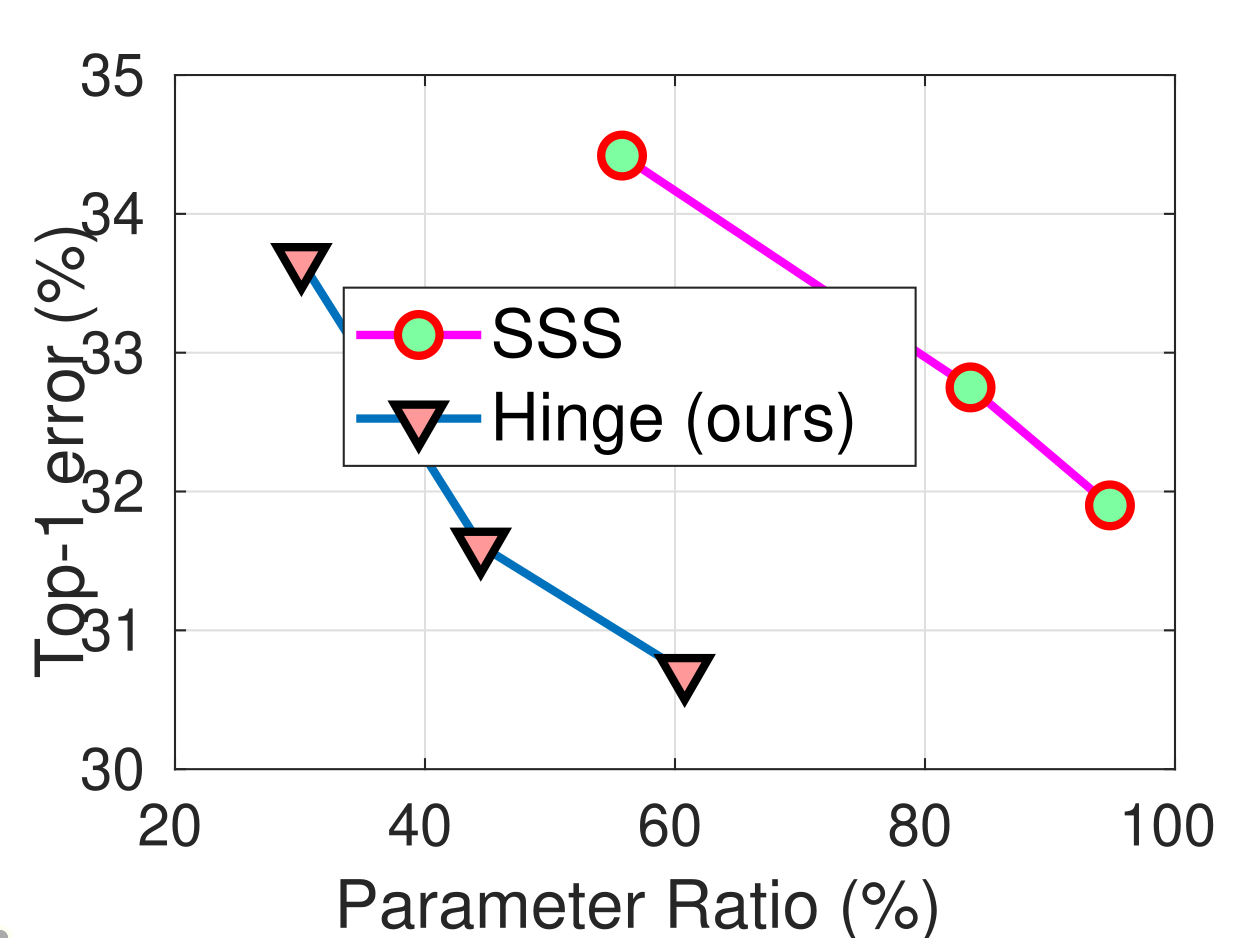}
  \subcaption{CIFAR100, ResNet20}
  \vspace{-0.2cm}
  \label{fig:cifar100_resnet20}
\end{minipage}%
\begin{minipage}[c]{.25\textwidth}
  \vspace*{\fill}
  \centering
  \includegraphics[width=1.04\linewidth]{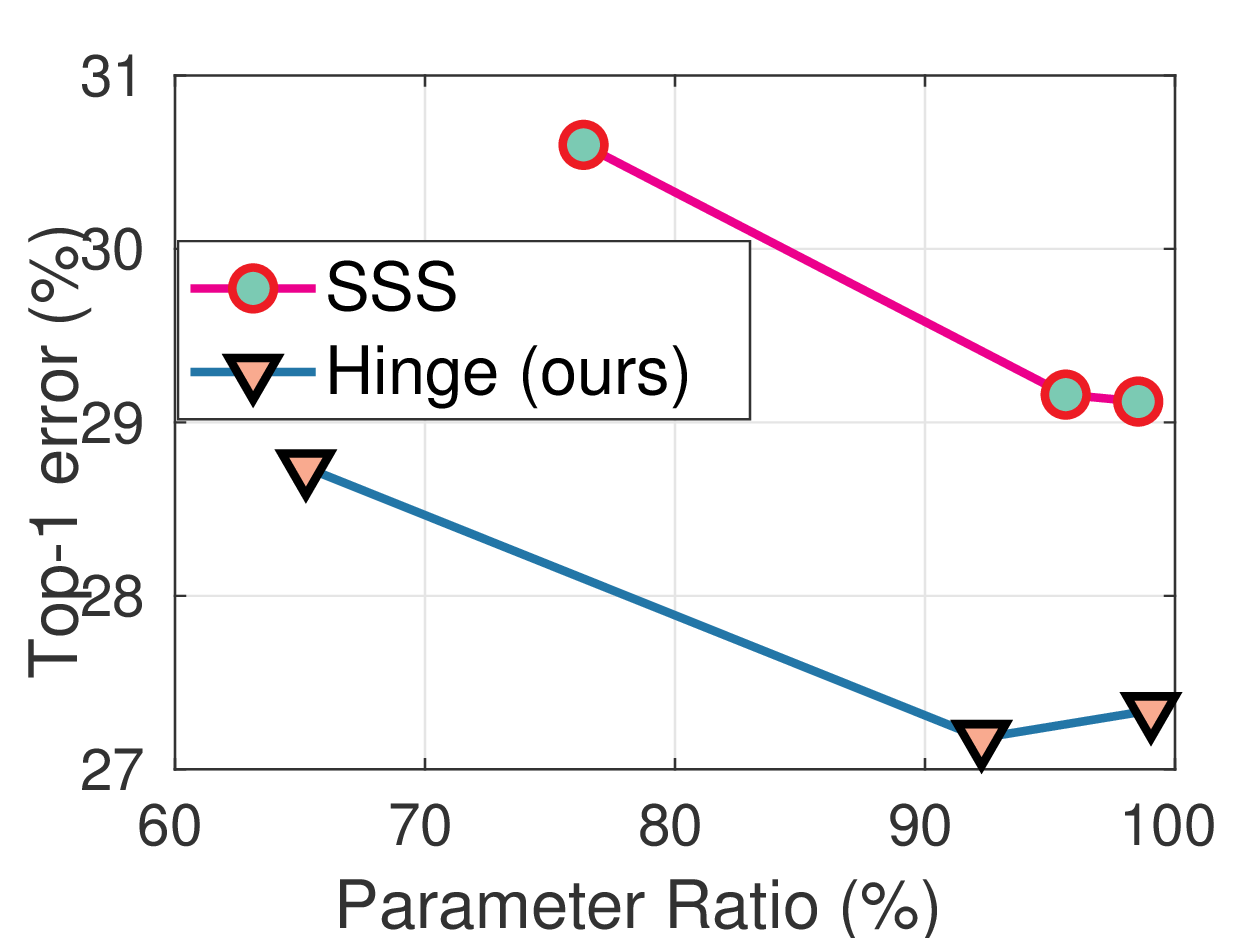}
  \subcaption{CIFAR100, ResNeXt20}
  \vspace{-0.2cm}
  \label{fig:cifar100_resnext20}
\end{minipage}%
\caption{Comparison between SSS~\cite{huang2018data} and the proposed method. Top-1 error rate is reported. (a) and (b) shows the results on CIFAR10 while (c) and (d) shows the results on CIFAR100.}
    \vspace{-0.4cm}
\label{fig:sss_hinge_comparison_params}
\end{figure*}

\section{Hyper Parameters for Different Regularizers}
\label{eqn:hyper_parameters}

The regularization factors for different regularizers are listed in Table~\ref{tbl:regularization_factor}. For CIFAR10 and CIFAR100 datasets, the learning rate $\eta$ of the sparsity-inducing matrix $\mathbf{A}$ during compression optimization is set to 0.1. The ratio between the learning rate of $\mathbf{W}$ and $\mathbf{A}$ is set to 0.01. That is, the learning rate $\eta_s$ of $\mathbf{W}$ during compression optimization is 0.001. For ImageNet, both $\eta$ and $\eta_s$ during optimization are set to 0.001.

\section{More Parameter Comparison}

In Fig.~\ref{fig:layer164_cifar100_params} and Fig.~\ref{fig:sss_hinge_comparison_params}, more parameter comparison results are shown. The figures report several operating points of the proposed method and SSS~\cite{huang2018data}. The proposed method forms a lower error bound for SSS. In Fig.~\ref{fig:layer164_cifar100_params}, our Hinge method without distillation loss is already better than SSS. And with the distillation loss, the proposed method shoots even lower Top-1 error rate.

\section{Layer-wise Compression Ratio}

The layer-wise or block-wise compression ratio of the model compressed by the proposed method is shown in Fig.~\ref{fig:ratio1} and Fig.~\ref{fig:ratio2}, respectively. For ResNeXt~\cite{xie2017aggregated}, the aim is to compress the $3 \times 3$ convolution in the residual block and the two $1 \times 1$ convolutions are used as the sparsity-inducing matrices. Thus, the block-wise compression ratio is reported. For ResNet~\cite{he2016deep} and WRN~\cite{zagoruyko2016wide}, there are two $3 \times 3$ convolutions in each residual block. Each of the two convolutions is compressed by introducing a sparsity-inducing matrix. Thus, the layer-wise compression ratio is reported. As shown in the Fig.~\ref{fig:ratio_wrn}, Fig.~\ref{fig:ratio_resnext164} and Fig.~\ref{fig:ratio_resnet164}, for WRN, ResNeXt164, and ResNet164, our approach tends to compress the shallow layers more compared with the deep layers. This is consistent with former research~\cite{he2017channel}. As for ResNet56 in Fig~\ref{fig:ratio_resnet56}, the proposed method results in a sawtooth architecture. That is, for the convolutions with the same feature dimension (\ie Layer 1 to Layer 18, Layer 19 to Layer 36, and Layer 37 to Layer54), the middle layers generally have a severer degree of compression.

\end{document}